\documentclass[onecolumn,final]{elsart}
\usepackage[english]{babel}
\usepackage{graphics}
\usepackage{array}
\usepackage{vmargin}
\newcommand{\bb}[1]{\mbox{\boldmath$#1$}}
\newcommand{\tild}{\char126}
\begin{document}
\begin{frontmatter}
\title{Parametric stiffness analysis of the Orthoglide}
\author{F. Majou$^{1,2}$, C. Gosselin$^2$, P. Wenger$^1$, D. Chablat$^{1,\bullet}$} 
\address{$^1$ Institut de Recherches en Communications et Cybern\'etique de Nantes}
\address{UMR CNRS 6597, 1 rue de la Noë, 44321 Nantes, France}
\address{$^2$  D\'epartement de G\'enie M\'ecanique, Universit\'e, Laval, Qu\'ebec, Canada, G1K 7P4}
\address{$\bullet$Corresponding author. Tel.: +33-2-40-37-69-48; fax: +33-2-40-37-69-30 {\it{E-mail address:}} Damien.Chablat@irccyn.ec-nantes.fr}
%%%%%%%%%%%%%%%%%%%%%%%%%%%%%%%%%%%%%%%%%%
%%%%%%%%%%%%%%%%%%%%%%%%%%%%%%%%%%%%%%%%%%
%\newpage
\selectlanguage{english}
\maketitle
\begin{abstract}
This paper presents a parametric stiffness analysis of the Orthoglide. A compliant modeling and a symbolic expression of the stiffness matrix are conducted. This allows a simple systematic analysis of the influence of the geometric design parameters and to quickly identify the critical link parameters. Our symbolic model is used to display the stiffest areas of the workspace for a specific machining task. Our approach can be applied to any parallel manipulator for which stiffness is a critical issue.
\end{abstract}
%%%%%%%%%%%%%%%%%%%%%%%%%%%%%%%%%%%%%%%%%%
\begin{keyword}
Parametric analysis; Stiffness; PKM design; Orhtoglide.
\end{keyword}
\end{frontmatter}
%%%%%%%%%%%%%%%%%%%%%%%%%%%%%%%%%%%%%%%%%%
\section{Introduction}
%%%%%%%%%%%%%%%%%%%%%%%%%%%%%%%%%%%%%%%%%%
Usually, parallel manipulators are claimed to offer good stiffness and accuracy properties, as well as good dynamic performances. This makes them attractive for innovative machine-tool structures for high speed machining \cite{Tlusty99,Wenger99,Majou01}. When a parallel manipulator is intended to become a Parallel Kinematic Machine (PKM), stiffness becomes a very important issue in its design \cite{Pritschow97,Company02,Brogardh02}. This paper presents a parametric stiffness analysis of the Orthoglide, a 3-axis translational PKM prototype developed at IRCCyN \cite{Wenger00}.

Finite Element Methods (FEM) are mandatory to carry out the final design of a PKM \cite{Bouzgarrou04}. However, a comprehensive three-dimensional FEM analysis may prove difficult, since one must repeatedly re-mesh the PKM structure to determine stiffness performances in the whole workspace, which is time consuming. Simpler and faster methods are needed at a pre-design stage. One of the first efficient stiffness analysis methods for parallel mechanisms was based on a kinetostatic modeling \cite{Gosselin90}. According to this approach, the stiffness of parallel mechanisms is mapped onto their workspace by taking into account the compliance of the actuated joints only. It is used and complemented in \cite{digregorio99} to show the influence of the compliance of the prismatic joints as well as the torsional compliance of the links on the stiffness of the 3-{\it UPU} mechanism assembled for translation \cite{Tsai00}. It is shown that the compliance of the links reduces the kinetostatic performances in a large part of the workspace, compared to the stiffness model based on rigid links. Furthermore, the mobile platform can undergo small rotational motions because of the links' compliance, which departs from the expected translational kinematic behavior.

The analysis presented in \cite{Gosselin90} is not appropriate for PKM whose legs, unlike hexapods, are subject to bending \cite{Kong02}. This problem is solved in \cite{Huang02}, where a stiffness estimation of a tripod-based overconstrained PKM is proposed. According to this approach, the PKM structure is decomposed into two substructures, one for the mechanism and another for the frame. One stiffness model is derived for each substructure. The superposition principle allows one to join the two models in order to derive the stiffness model of the whole structure. The influence of the geometrical parameters on the stiffness is also briefly studied. An interesting aspect of this method is that it can deal with overconstrained structures. However this stiffness model is not general enough. A more general model was proposed in \cite{Gosselin02}. The method is based on a flexible-link lumped parameter model that replaces the compliance of the links by localized virtual joints and rigid links. The latter approach differs from that presented in \cite{Huang02} on two main points, namely: $(i)$ the modeling of the link compliances and $(ii)$ the more general nature of the equations allowing the computation of the stiffness model.

In this paper, the method proposed in \cite{Gosselin02} is applied to the Orthoglide for a parametric stiffness analysis. A symbolic expression of the stiffness matrix is obtained which allows a global analysis of the influence of the Orthoglide's critical design parameters. No numerical computations are conducted until graphical results are generated. This paper is organized as follows: first the Orthoglide is presented. Then, the compliant model is introduced and the stiffness model is computed. Analytical expressions of the components of the stiffness matrix are obtained at the isotropic configuration, clearly showing the influence of each geometrical parameter. Finally, given a specific simulated machining task, it is shown how the general stiffness expressions allow one to easily display the stiffest subvolume of the Orthoglide's workspace.
%%%%%%%%%%%%%%%%%%%%%%%%%%%%%%%%%%%%%%%%%%
\section{Compliant modeling of the Orthoglide}
\label{Ortho}
%%%%%%%%%%%%%%%%%%%%%%%%%%%%%%%%%%%%%%%%%%
%%%%%%%%%%%%%%%%%%%%%%%%%%%%%%%%%%%%%%%%%%
\subsection{Kinematic architecture of the Orthoglide}
\begin{figure}
  \begin{centering}
    \begin{tabular}{c c}
       \begin{minipage}[t]{80 mm}
        {\scalebox{1}
        {\includegraphics{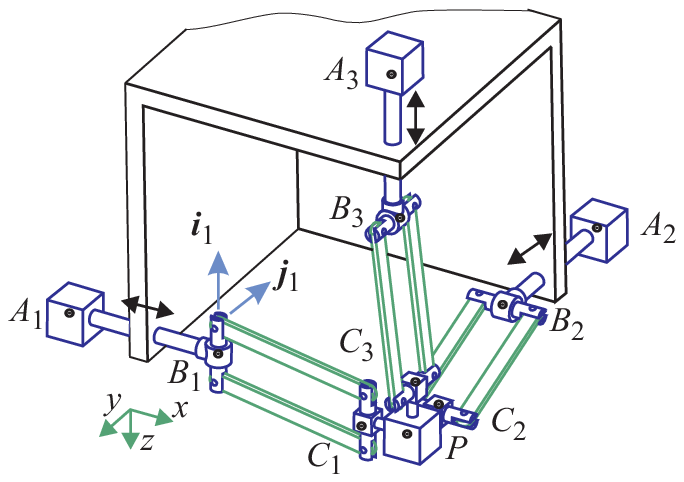}}}(a)
       \end{minipage} & 
       \begin{minipage}[t]{60 mm}
        {\scalebox{.35}
        {\includegraphics{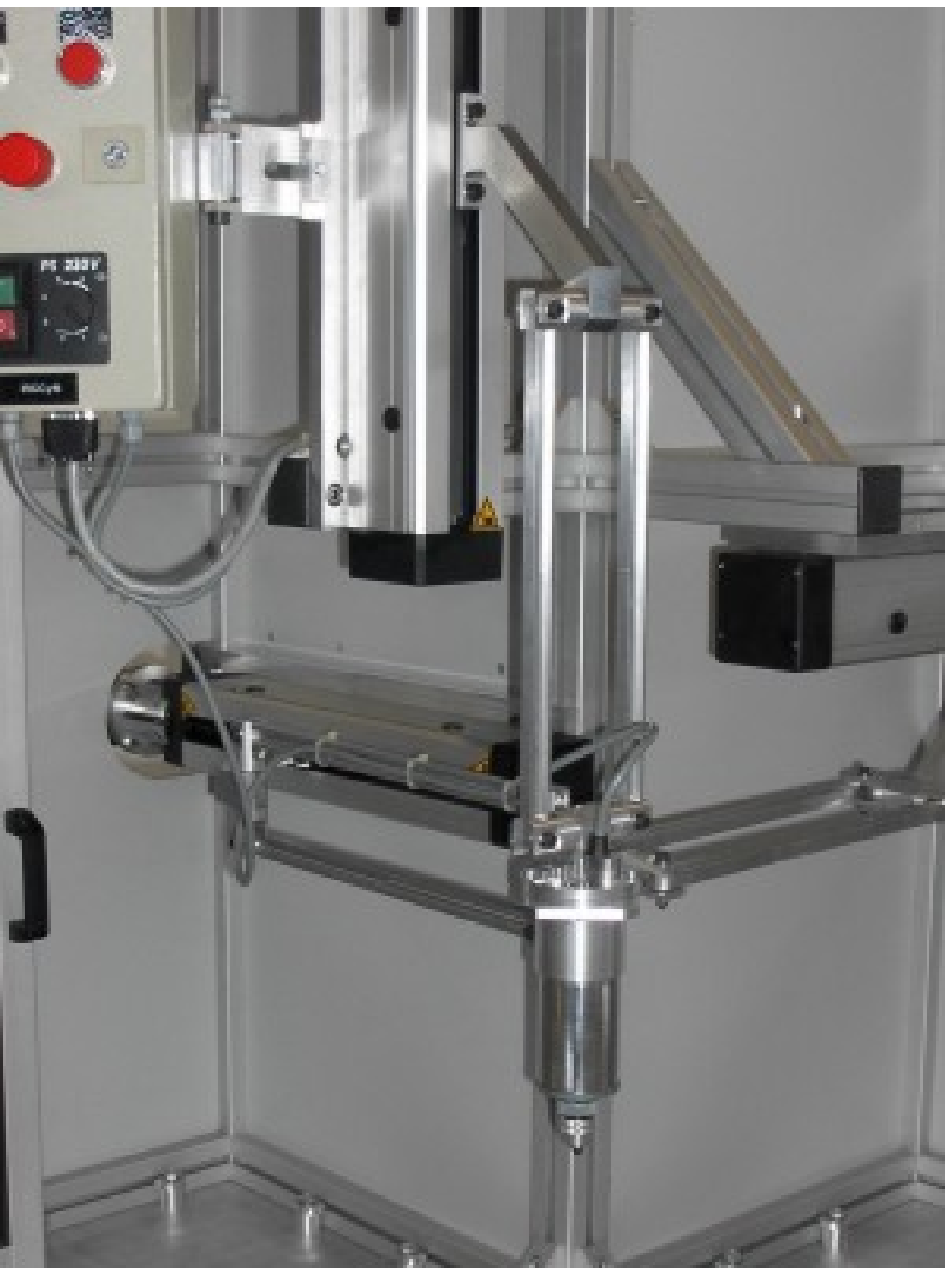}}(b)}
       \end{minipage}
     \end{tabular}
\caption{The Orthoglide (a) Kinematic architecture and (b) Prototype}
\label{ortho_cine}
\end{centering}
\end{figure}

%%%%%%%%%%%%%%%%%%%%%%%%%%%%%%%%%%%%%%%%%%
The Orthoglide is a translational 3-axis PKM prototype designed for machining applications. The mobile platform is connected to three orthogonal linear drives through three identical $RP_aR$ serial chains (Fig.\ref{ortho_cine}). Here, $R$ stands for a revolute joint and $P_a$ for a parallelogram-based joint. The Orthoglide moves in the Cartesian workspace while maintaining a fixed orientation. The Orthoglide was optimized for a prescribed workspace with prescribed kinetostatic performances \cite{Chablat03}. Its kinematic analysis, design and optimization are fully described in \cite{Chablat03}.
%%%%%%%%%%%%%%%%%%%%%%%%%%%%%%%%%%%%%%%%%%
\subsection{Parameters for compliant modeling}
\begin{figure}[hb!]
  \begin{center}
    \begin{tabular}{c c}
       \begin{minipage}[t]{70 mm}
         \begin{center}
         \resizebox{!}{3.2cm}
         {\includegraphics{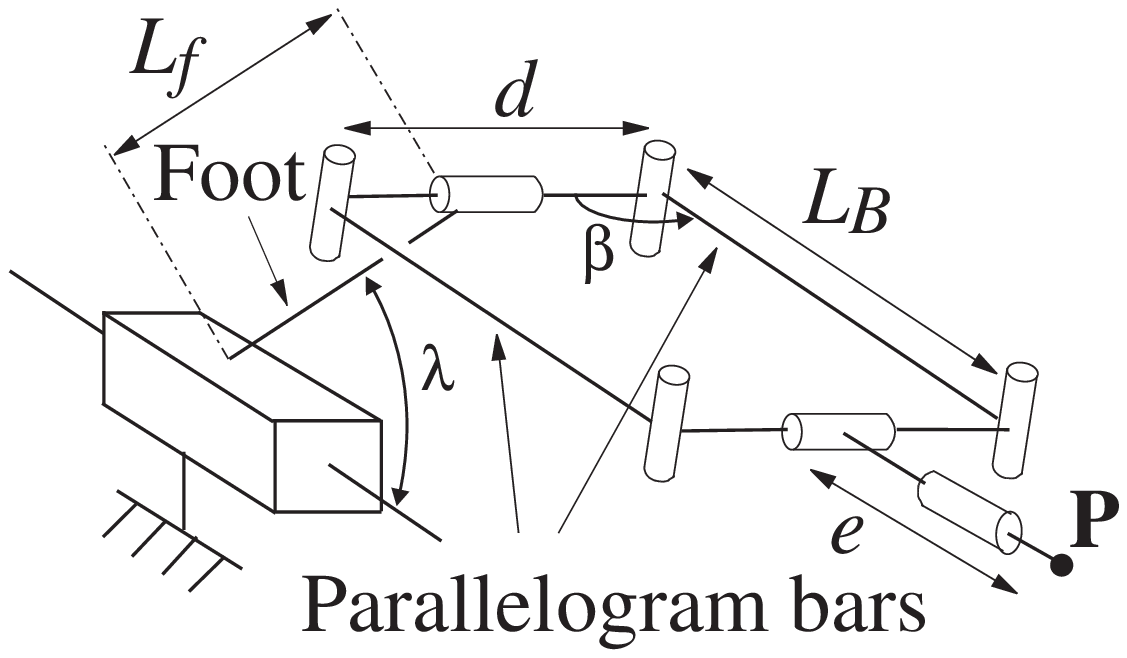}}
         \caption{Geometric parameters of the leg}
         \label{leg_parameters}
         \end{center}
       \end{minipage} &
       \begin{minipage}[t]{70 mm}
         \begin{center}
         {\scalebox{.45}
        {\includegraphics{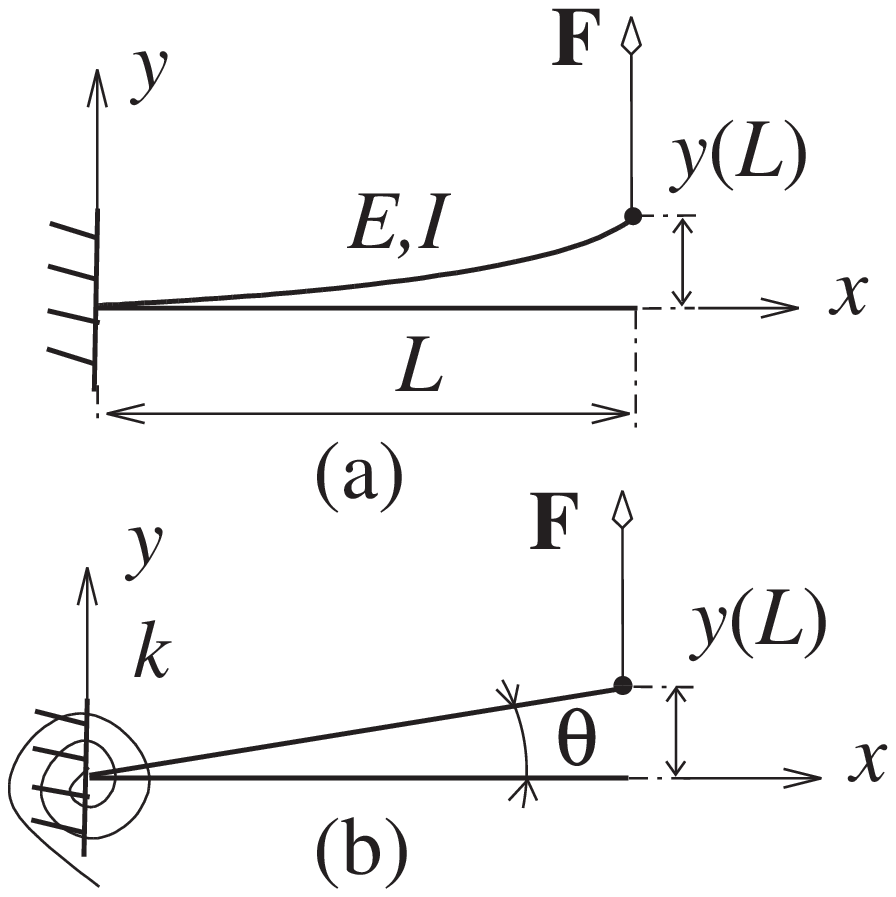}}}
         \caption{General model for a flexible link 
         (a) Flexible beam 
         (b) Virtual rigid beam}
         \label{poutre_flex}
         \end{center}
       \end{minipage} 
    \end{tabular}
  \end{center}
\end{figure}

%%%%%%%%%%%%%%%%%%%%%%%%%%%%%%%%%%%%%%%%%%
The parameters used for the compliant modeling of the Orthoglide are presented on Fig.~\ref{leg_parameters} and in Tab.~\ref{parameters}. They correspond to a ``beam-like'' modeling of the Orthoglide legs' links. The foot has been designed to prevent each parallelogram from colliding with the corresponding linear motion guide.
Three revolute joints are added, one on each leg (see Fig. \ref{leg_parameters}), because the stiffness method used does not work with an overconstrained Orthoglide. This does not change the kinematics.
\begin{table}[hb!]
  \begin{center}
  {\begin{tabular}[t]{|c|p{8cm}|p{1.5cm}|}
  \hline {Parameter} & {Description} & Values\\ \hline
  $L_f$ & Foot length, see Fig.\ref{leg_parameters} & 150mm\\
  $h_f$ &  Foot section sides & 26mm\\
  $b_f$ &  Foot section sides & 16mm\\
  $I_{f_1}=\frac{b_f.h_f^3}{12}$ &  Foot section moment of inertia 1&\\
  $I_{f_2}=\frac{h_f.b_f^3}{12}$  & Foot section moment of inertia 2&\\
  $I_{f_0}=h_f.b_f(h_f^2+b_f^2)/12$& Foot section polar quadratic moment &\\
  $\lambda$ & Angle between foot axis and actuated joint axis, see Fig.\ref{leg_parameters}& $45^{\circ}$\\
  $d$  &  Distance between parallelogram bars, see Fig.\ref{leg_parameters} & 80mm\\
  $L_B$ & Parallelogram bar length, see Fig.\ref{leg_parameters} & 310mm\\
  $S_B$ & Parallelogram bar cross-section area & $144mm^2$\\
  $\beta$& Rotation angle of the parallelogram&\\
  $e$ &  See Fig.\ref{leg_parameters}&\\ \hline
\end{tabular}}
\caption{Geometric parameters of the Orthoglide and dimensions of the protptype} \label{parameters}
\end{center}
\end{table}
%%%%%%%%%%%%%%%%%%%%%%%%%%%%%%%%%%%%%%%%%%
\subsection{Compliant modeling with flexible links}
\label{compliance}
\begin{figure}[ht!]
    \begin{center}
           \scalebox{.45}{\includegraphics{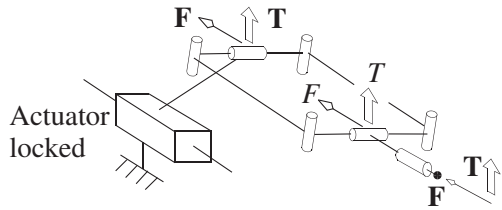}}
           \caption{Forces transmitted in a leg}
           \label{distri_efforts}
    \end{center}
\end{figure}

%%%%%%%%%%%%%%%%%%%%%%%%%%%%%%%%%%%%%%%%%%
In the lumped model described in \cite{Gosselin02}, the leg links are considered as flexible beams and are  replaced by rigid beams mounted on revolute joints plus torsional springs located at the joints (Fig. \ref{poutre_flex}).
Deriving the relationship between the force ${\bf F}$ and the deformation $y(x)$, the local torsional stiffness $k$ can be computed:
\begin{eqnarray}
   {EIy''(x)} & {=} & {F(L-x)} \nonumber \\ 
   {~} & {\vdots} & {~} \nonumber \\ 
   {EIy(L)} & {=} & {FL^3/3} \nonumber \\
   {\rightarrow \theta \simeq y(L)/L} & {=} & {FL^2/3EI}  \nonumber \\ 
   {k}  &  {=}  & {FL/\theta}   \nonumber \\  
   {\rightarrow k}  &  {=}  & {3EI/L}   \nonumber
\end{eqnarray}
If the Orthoglide leg actuator is locked, then one leg can withstand one force ${\bf F}$ and one torque ${\bf T}$ (Fig. \ref{distri_efforts}), which are transmitted along the parallelogram bars and the foot. For a compliant modeling that uses virtual joints, it is important to understand how external forces are transmitted, and what their effect on the leg links is.
Eight virtual joints are modeled along the Orthoglide leg. They are described in Tab. \ref{tab_flex_first}. The determination of all the virtual joint stiffnesses is not detailed here for brevity. However, they are derived based on the same principles used to calculate the torsional stiffness above.

The actuated joint is assumed to be much stiffer than the virtual joints. The leg links compliances modeled in Tab. \ref{tab_flex_first} were selected beforehand as the most significant ones. Indeed, selecting only the most significant compliances plays an important role in reducing the computing time required to derive the stiffness matrix symbolically (Par. \ref{stiff_model}). The kinematic joints' compliances are not taken into account because our purpose is to determine the links compliance influence only. Angle $\beta$ is a parameter that depends on the Cartesian coordinates.
%--------------------TABLEAU-----------------------
\begin{table}[b!]
\begin{center}
%   \vspace{2mm}%\small
  {\begin{tabular}{|b{2.6cm}|c||b{3.3cm}|c|} \hline
  {\it Virtual joints $i$} & 
  {\it Figure} & 
  {\it Virtual joints $i$} & 
  {\it Figure} \\  \hline
  {$k_1=k_{act}$
  \newline translational 
  \newline stiffness of the 
  \newline prismatic 
  \newline actuator} &
  {\scalebox{.45}
  {\includegraphics{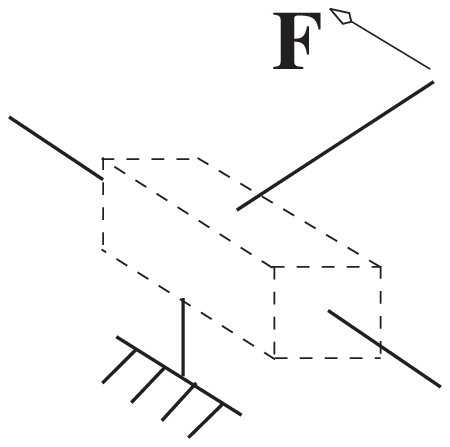}}} &
  {$k_5=\frac{EI_{f_2}}{L_f}$ 
  \newline Foot section 
  \newline rotation due to 
  \newline torque $\bf T$} &
  {\scalebox{.45}
  {\includegraphics{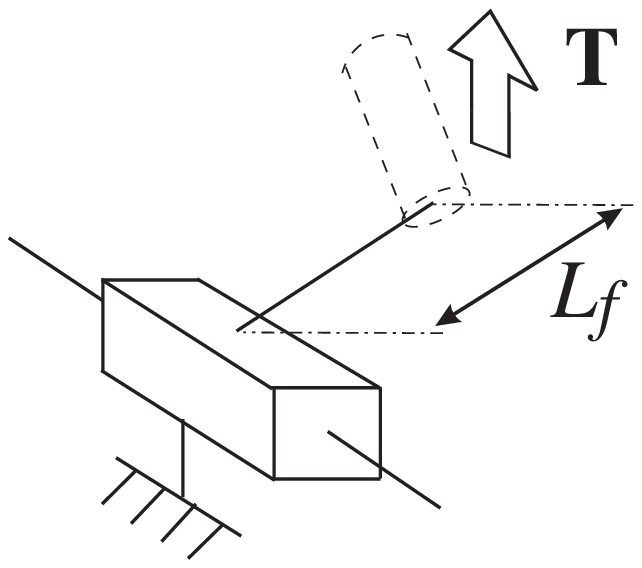}}} \\
  \hline
  {$k_2=\frac{3EI_{f_1}}{L_f}$ 
  \newline Foot bending 
  \newline due to force $\bf F$} &
  {\scalebox{.45}
  {\includegraphics{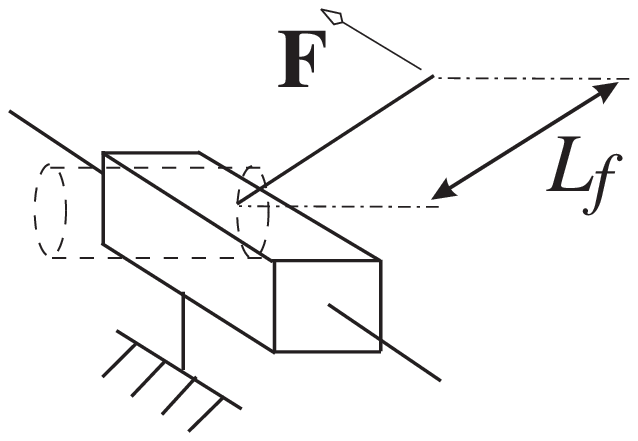}}}& 
  {$k_8=\frac{2ES_B}{L_B}$
  \newline Parallelogram 
  \newline bars tension/
  \newline compression due 
  \newline to force $\bf F$} &
  {\scalebox{.45}
  {\includegraphics{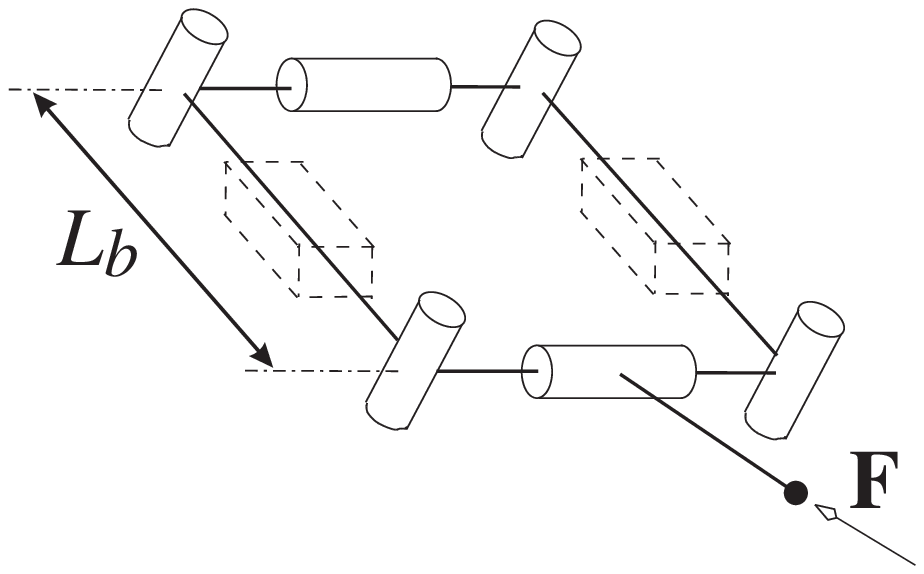}}} \\
  \hline
  {$k_3=\frac{2EI_{f_2}}{L_f}$
  \newline Foot bending 
  \newline due to torque $\bf T$} &
  {\scalebox{.45}
  {\includegraphics{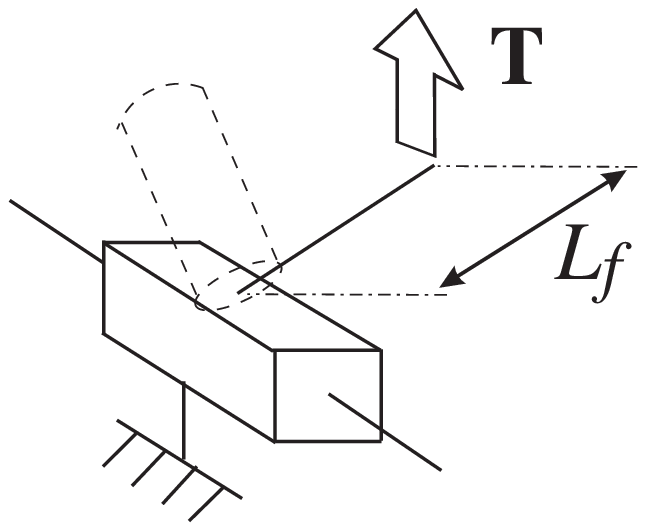}}} & 
  {$k_{10}=\frac{ES_Bd^2\cos(\beta)}{2L_B}$
  \newline Differential 
  \newline tension of 
  \newline parallelogram bars 
  \newline due to torque $\bf T$} &
  {\scalebox{.45}
  {\includegraphics{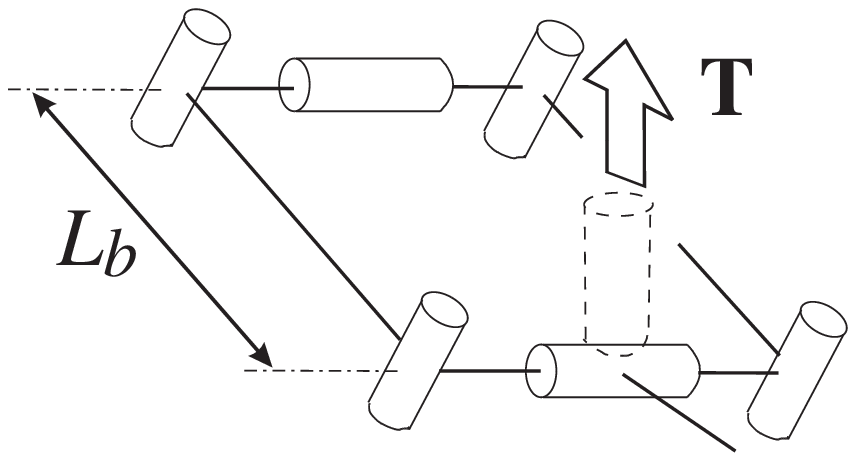}}} \\
  \hline
  {$k_4=\frac{GI_{f_O}}{L_f}$
  \newline Foot torsion  
  \newline due to torque $\bf T$} &
  {\scalebox{.45}
  {\includegraphics{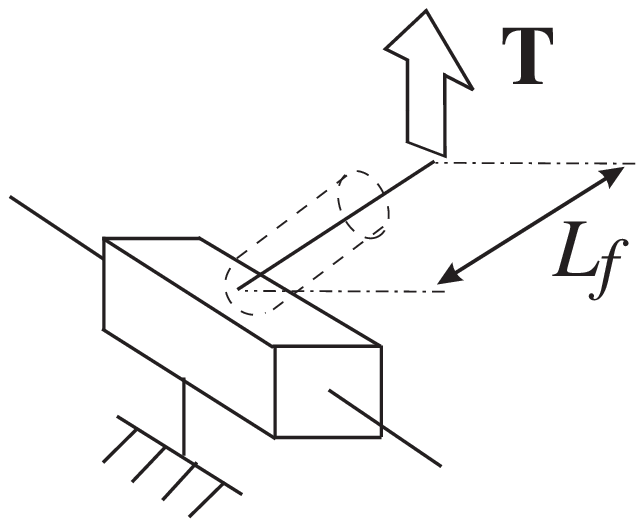}}}\\
  \cline{1-2}
  \end{tabular}}
    \caption{Virtual joints modeling}
    \label{tab_flex_first}
\end{center}
\end{table}
%%%%%%%%%%%%%%%%%%%%%%%%%%%%%%%%%%%%%%%
\section{Symbolic derivation of the stiffness matrix}
\label{stiff_model}
\begin{figure}[ht!]
\begin{centering}
        {\scalebox{.45}{\includegraphics{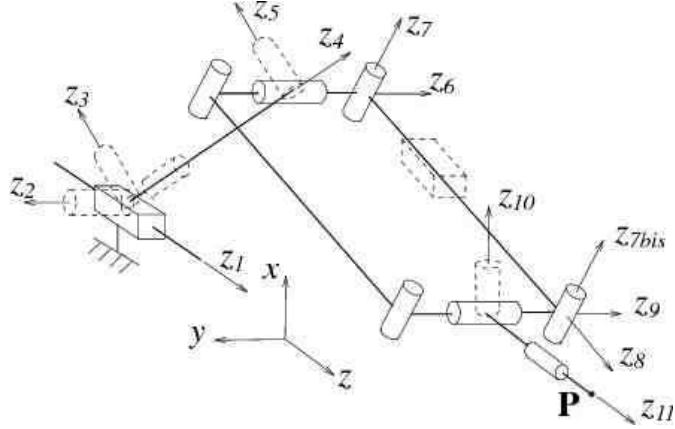}}}
        \caption{Flexible leg}
        \label{jambe_ortho_flex}
\end{centering}
\end{figure}

%%%%%%%%%%%%%%%%%%%%%%%%%%%%%%%%%%%%%%%
In this section, the derivation of the Orthoglide stiffness matrix --- based on the virtual joints described in the previous section ---  is conducted with a stiffness model that was fully described in \cite{Gosselin02}. Therefore, the description of the model will only be summarized here. Fig. \ref{jambe_ortho_flex} represents the lumped model of a leg with flexible links.
The Jacobian matrix ${\bf{J}}_i$ of the $i$th leg of the Orthoglide is obtained from the Denavit-Hartenberg parameters of the $i$th leg with flexible links. This matrix maps all leg joint rates (including the virtual joints) into the generalized velocity of the platform, i.e.,
$$
  {\bf J}_i {\dot {\bb \theta}}_i = {\bf t} \quad
{\rm where} \quad
{\dot{\bb{\theta}}}_i^T =
  [\begin{array}{ccccccccccc} 
    {\dot{\theta}}_{i_1} &
    {\dot{\theta}}_{i_2} & 
    {\dot{\theta}}_{i_3} & 
    {\dot{\theta}}_{i_4} & 
    {\dot{\theta}}_{i_5} & 
    {\dot{\theta}}_{i_6} &
    {\dot{\theta}}_{i_7} & 
    {\dot{\theta}}_{i_8} & 
    {\dot{\theta}}_{i_9} & 
    {\dot{\theta}}_{i_{10}} & 
    {\dot{\theta}}_{i_{11}}
  \end{array}]
$$
is the vector containing the 11 actuated, passive and virtual joint rates of leg i and $\bf t$ is the twist of the platform. The $P_a$ joint parameterization imposes ${\dot{\theta}}_{i_7}= - {\dot{\theta}}_{i_{7bis}}$, which makes ${\dot{\theta}}_{i_7}$ and ${\dot{\theta}}_{i_{7bis}}$  dependent. ${\dot{\theta}}_{i_7}$ is chosen to model the circular translational motion, and finally ${\bf J}_i$ is written as
  \begin{eqnarray}
{\bf J}_i &=& \left[\begin{array}{ccccccc} 0, & {\bf e}_{i_2}, &
{\bf e}_{i_3}, & {\bf e}_{i_4}, & {\bf e}_{i_5}, & {\bf e}_{i_6} \\
{\bf e}_{i_1}, & {\bf e}_{i_2}\times{\bf r}_{i_2}, & {\bf
e}_{i_3}\times{\bf r}_{i_3}, & {\bf e}_{i_4}\times{\bf r}_{i_4}, &
{\bf e}_{i_5}\times{\bf r}_{i_5}, & {\bf e}_{i_6}\times{\bf
r}_{i_6},
\end{array}
\right. \nonumber \\
&&\left. \begin{array}{ccccc}
0, & 0, & {\bf e}_{i_9}, & {\bf e}_{i_{10}}, & {\bf e}_{i_{11}}\\
{\bf e}_{i_7}\times{\bf r}_{i_7}-{\bf e}_{i_{7bis}}\times{\bf
r}_{i_{7bis}}, & {\bf e}_{i_8}\times{\bf r}_{i_8}, & {\bf
e}_{i_9}\times{\bf r}_{i_9}, & {\bf e}_{i_{10}}\times{\bf
r}_{i_{10}}, & {\bf e}_{i_{11}}\times{\bf r}_{i_{11}}
                    \end{array}\right] \nonumber
  \end{eqnarray}
%\end{flushright}
%%%%%%%%%%%%%%%%%%%%%%%%%%%%%%%%%%%%%%%
\noindent  in which ${\bf e}_{i_j}$ is the unit vector along joint j of leg i and ${\bf r}_{i_j}$ is the vector connecting joint j of leg i to the platform reference point. Therefore the Jacobian matrix of the Orthoglide can be written as:
$$
{\bf J} = \left[\begin{array}{ccc}
                 {\bf J}_1 & {\bf 0}   & {\bf 0}    \\
                 {\bf 0}   & {\bf J}_2 & {\bf 0}    \\
                 {\bf 0}   & {\bf 0}   & {\bf J}_3  \\
                \end{array}\right]
$$
One then has:
\begin{equation}
  \label{eq_J}
  {\bf J}\dot{\bb{\theta}} = {\bf R}{\bf t}
  \quad 
  {\rm with} 
  \quad 
  {\bf R}=[{\bf I}_6~{\bf I}_6~{\bf I}_6]^T
  \quad 
  {\rm and} 
  \quad 
  {\bf t} = \left\{\begin{array}{c} {\bb \omega} \\
  {\bf v}
  \end{array}\right\}
\end{equation}
\noindent $\dot{\bb{\theta}}$ being the vector of the 33 joint rates, that is ${\dot{\bb{\theta}}} = [{\dot{\bb{\theta}}}_1^T ~ {\dot{\bb{\theta}}}_2^T ~{\dot{\bb{\theta}}}_3^T]^T$. {\bf I}$_6$ stands for the $6\times 6$ identity matrix. Unactuated joints are then eliminated by writing the geometric conditions that constrain the two independent closed-loop kinematic chains of the Orthoglide kinematic structure:
\begin{equation}
  \label{closchain} 
  {\bf J}_1{\dot{\bb \theta}}_1 = 
  {\bf J}_2{\dot{\bb \theta}}_2
  \quad {\rm and} \quad
  {\bf J}_1{\dot{\bb \theta}}_1 = 
  {\bf J}_3{\dot{\bb \theta}}_3
\end{equation}
\noindent From (\ref{closchain}), one can obtain ${\bf A}{\dot {\bb \theta'}}= {\bf B}{\dot {\bb \theta''}}$
(see \cite{Gosselin02} for details), where $\dot {\bb \theta}'$ is the vector of joint rates without passive joints and  $\dot {\bb \theta}''$ is the vector of joint rates with only passive joints. Hence: 
$$
  \dot {\bb \theta}'' = {\bf B}^{-1}{\bf A} \dot {\bb \theta}' 
$$ 
Then a matrix $\bf{V}$ is obtained (see \cite{Gosselin02} for details) such that:
 \begin{equation}
  \label{eq_V} 
  \dot {\bb \theta}={\bf V} \dot {\bb \theta}'
  \end{equation}
\noindent From (\ref{eq_J}) and (\ref{eq_V}) one can obtain:
\begin{equation}
  \label{eq_JV}
  {\bf J}{\bf V}\dot{\bb \theta}' = {\bf R}{\bf t}
\end{equation}
\noindent As matrix ${\bf R}$ represents a system of 18 compatible linear equations in 6 unknowns, one can use the least-square solution to obtain an exact solution from (\ref{eq_JV}):
 $$
 {\bf t} = ({\bf R}^T{\bf R})^{-1}{\bf R}^T{\bf J}{\bf V}\dot{\bb \theta}'
 $$
Now let  ${\bf J}'$ be represented as ${\bf J}' = ({\bf R}^T{\bf R})^{-1}{\bf R}^T{\bf J}{\bf V}$. Then one has:
\begin{equation}
  \label{eq_t} 
  {\bf t} = {\bf J}'\dot{\bb \theta}'
\end{equation}
According to the principle of virtual work, one has:
\begin{equation}
  \label{pvw} 
  {\bb \tau}^T {\dot {\bb \theta}'} = {\bf w}^T{\bf t}
\end{equation}
\noindent where ${\bb \tau}$ is the vector of forces and torques applied at each actuated or virtual joint and $\bf w$ is the external wrench  applied at the end effector, point {\bf P}. Gravitational forces are neglected. By substituting (\ref{eq_t}) in (\ref{pvw}), one can obtain:
\begin{equation}
  \label{eq_tau}
  {\bb \tau} = {\bf J}'^T {\bf w}
\end{equation}
\noindent The forces and displacements of each actuated or virtual joint can be related by Hooke's law, that is for the whole structure one has:
%%%%%%%%%%%%%%%%%%%%%%%%%%%%%%%%%%%%%%%
\begin{equation}
\label{hooke} 
{\bb \tau} =  {\bf K}_J \Delta {\bb \theta'}
\end{equation}
\par
${\rm with} \quad
{\bf K}_J \! = \!
                  \left[\begin{array}{ccc}
                 {\bf A} & {\bf 0} & {\bf 0} \\
                 {\bf 0} & {\bf A} & {\bf 0} \\
                 {\bf 0} & {\bf 0} & {\bf A} \\
                  \end{array}\right] 
$
\par
${\rm and~} 
{\bf A}= {\rm diag}\left(
                \begin{array}{ccccccccc}
                  k_{act}, & 
                  \frac{3EI_{f_1}}{L_f}, & 
                  \frac{2EI_{f_2}}{L_f}, & 
                  \frac{GI_{f_O}}{L_f}, & 
                  \frac{Eh_fb_f}{L_f}, & 
                  \frac{EI_{f_2}}{L_f}, & 
                  \frac{2ES_B}{L_B}, &
                  \frac{ES_Bd^2\cos(\beta)}{L_B} 
                \end{array}
             \right).
$
\par
%%%%%%%%%%%%%%%%%%%%%%%%%%%%%%%%%%%%%%%
%\begin{equation}
%\label{matA}
%{\rm and} \quad
%{\bf A} = \left[\begin{array}{ccccccccc}
%                  k_{act} & 0 & 0 & 0 & 0 & 0 & 0 & 0\\
%                  0 & \frac{3EI_{f_1}}{L_f} & 0 & 0 & 0 & 0 & 0 & 0\\
%                  0 & 0 & \frac{2EI_{f_2}}{L_f} & 0 & 0 & 0 & 0 & 0\\
%                  0 & 0 & 0 & \frac{GI_{f_O}}{L_f} & 0 & 0 & 0 & 0\\
%                  0 & 0 & 0 & 0 & \frac{Eh_fb_f}{L_f} & 0 & 0 & 0 \\
%                  0 & 0 & 0 & 0 & 0 & \frac{EI_{f_2}}{L_f} & 0 & 0 \\
%                  0 & 0 & 0 & 0 & 0 & 0 & \frac{2ES_B}{L_B} & 0 \\
%                  0 & 0 & 0 & 0 & 0 & 0 & 0 & \frac{ES_Bd^2\cos(\beta)}{L_B} \\
%                  \end{array}\right]
%\end{equation}
%%%%%%%%%%%%%%%%%%%%%%%%%%%%%%%%%%%%%%%
$\Delta {\bb \theta'}$ only includes the actuated and virtual joints, that is by equating (\ref{eq_tau}) with (\ref{hooke}):
$$
  {\bf K}_J \Delta {\bb \theta'}={\bf J}'^T{\bf w}
$$  
\noindent Hence $\Delta {\bb \theta'}={\bf K}_J^{-1}{\bf J}'^T{\bf w}$. Pre-multiplying both sides by ${\bf J}'$ one obtains:
\begin{equation}
\label{J_delta}
 {\bf J}'\Delta {\bb \theta'}={\bf J}'{\bf K}_J^{-1}{\bf J}'^T{\bf w}
\end{equation}
\noindent Substituting (\ref{eq_t}) into (\ref{J_delta}), one obtains:
  $$
%  {\bf t}={\bf J}'{\bf K}_J^{-1}{\bf J}'^T{\bf w}
  {\bf d}={\bf J}'{\bf K}_J^{-1}{\bf J}'^T{\bf w}
  $$
  %%% CD
\noindent with ${\bf d}={\bf t} \Delta t$. Finally the compliance matrix $\bb\kappa$ is obtained as follows:
$$
  {\bb \kappa} ={\bf J}'{{\bf K}_J}^{-1}{\bf J}'^T
$$
In the Orthoglide case we obtain:
\begin{equation}
  \label{stiffmat} 
  {\bb \kappa} = \left(\begin{array}{cccccc}
                  {\kappa}_{11} & 
                  0 & 
                  0  & 
                  {\kappa}_{14} & 
                  {\kappa}_{15} & 
                  {\kappa}_{16}\\
                  0 & 
                  {\kappa}_{11} & 
                  0 & 
                  {\kappa}_{24} & 
                  {\kappa}_{25} & 
                  {\kappa}_{26}  \\
                  0 & 0 & 
                  {\kappa}_{11} & 
                  {\kappa}_{34} & 
                  {\kappa}_{35} & 
                  {\kappa}_{36}  \\
                  {\kappa}_{14} & 
                  {\kappa}_{24} & 
                  {\kappa}_{34} & 
                  {\kappa}_{44} & 
                  {\kappa}_{45} &
                  {\kappa}_{46} \\
                  {\kappa}_{15} & 
                  {\kappa}_{25} & 
                  {\kappa}_{35} & 
                  {\kappa}_{45} & 
                  {\kappa}_{55} & 
                  {\kappa}_{56} \\
                  {\kappa}_{16} & 
                  {\kappa}_{26} & 
                  {\kappa}_{36} & 
                  {\kappa}_{46} & 
                  {\kappa}_{56} & 
                  {\kappa}_{66}
                  \end{array}\right)
\end{equation}
%%%%%%%%%%%%%%%%%%%%%%%%%%%%%%%%%%%%%%%
And the Cartesian stiffness matrix is:
$$
  {\bf K}  = 
  {\bb \kappa}^{-1} = {({\bf J}'{{\bf K}_J}^{-1}{\bf J}'^T)}^{-1}
$$
%%%%%%%%%%%%%%%%%%%%%%%%%%%%%%%%%%%%%%%%%%
\section{Parametric stiffness analysis at the isotropic configuration}
\label{results}
%%%%%%%%%%%%%%%%%%%%%%%%%%%%%%%%%%%%%%%%%%
In this section, we study the influence of the geometric parameters on the stiffness of the Orthoglide at the isotropic configuration, since this configuration provides a good evaluation of the overall performances \cite{Chablat03}. Another interest is that the stiffness matrix is then diagonal which makes it easier to analyze.
%%%%%%%%%%%%%%%%%%%%%%%%%%%%%%%%%%%%%%%%%%
\subsection{Simple symbolic expressions}
%%%%%%%%%%%%%%%%%%%%%%%%%%%%%%%%%%%%%%%%%%
At the isotropic configuration, $\bb \kappa$ is diagonal and the symbolic expressions of the components $\kappa_{ij}$ are simple. This is convenient because it is then possible to invert $\bb \kappa$ within a Maple worksheet and then analyze the symbolic expressions of the components of matrix ${\bf K}$. We have:
$$ 
{\bf K}= {\rm diag}\left(\begin{array}{cccccc}
                      K_a, & K_a, & K_a, & K_b, & K_b, & K_b
                    \end{array}\right)
$$
%$$
%   {\bf K}
%  = \left(\begin{array}{cccccc}
%                  K_a & 0 & 0 & 0 & 0 & 0 \\
%                  0 & K_a & 0 & 0 & 0 & 0 \\
%                  0 & 0 & K_a & 0 & 0 & 0 \\
%                  0 & 0 & 0 & K_b & 0 & 0 \\
%                  0 & 0 & 0 & 0 & K_b & 0 \\
%                  0 & 0 & 0 & 0 & 0 & K_b
%                  \end{array}\right)
%$$
\noindent where $K_a$ is the torsional stiffness and $K_b$ is the translational stiffness.
\begin{equation}
  \label{stiffmat_iso_tors}
  K_a=\frac{E}{
  \frac{2L_B}{S_Bd^2}+
  \frac{2L_p(78b_f^2+\cos^2\lambda(45h_f^2-33b_f^2))}{5h_fb_f^3(b_f^2+h_f^2)}}
  \quad
  K_b=\frac{1}{
  \frac{1}{k_{act}}+
  \frac{L_B}{2S_BE}+
  \frac{4L_f^3\sin^2\lambda}{Eh_f^3b_f}}
\end{equation}
Analyzing the Orthoglide's stiffness at the isotropic configuration allows us to manipulate simple and meaningful symbolic expressions that are easy to interpret: this is the purpose of the following subsections.
%%%%%%%%%%%%%%%%%%%%%%%%%%%%%%%%%%%%%%%%%%
\subsection{Qualitative analysis of $K_a$ and $K_b$}
%%%%%%%%%%%%%%%%%%%%%%%%%%%%%%%%%%%%%%%%%%
By inspection of the symbolic expression of $K_a$ a few observations can be made:
\begin{itemize}
  \item Young's modulus $E$ appears at the numerator, which makes its influence easy to understand: when $E$ increases, $K_a$ increases, which is in accordance with intuition;
  \item The term $\frac{2L_B}{S_Bd^2}$ shows the influence of virtual joint $10$ (differential tension of parallelogram bars). When the bar length $L_B$ increases or when $S_B$ decreases, $K_a$ decreases which is also in accordance with intuition. $K_a$ decreases when $d$ increases, which is a less intuitive result\footnote{Note that should $d$ increase above a certain limit, other links compliances previously ruled out as less significant may then need to be taken into account.};
  \item The expression $\frac{2L_p(78b_f^2+\cos^2\lambda(45h_f^2-33b_f^2))}{5h_fb_f^3(b_f^2+h_f^2)}$ shows the influence of virtual joints $3$, $4$ and $5$ (foot bending and torsion). $K_a$ decreases when $L_f$ increases, which is not surprising. The degrees of $h_f$ and $b_f$ in the numerator and denominator of $K_a$ tend to prove that the rotational stiffness increases with $h_f$ or $b_f$, which is in accordance with intuition. The influence of $\lambda$ depends on the sign of $(45h_f^2-33b_f^2)$.
\end{itemize}
\noindent Similarly, by inspection of the symbolic expression of $K_b$ one notes:
\begin{itemize}
  \item The term $\frac{1}{k_{act}}$ shows the influence of the prismatic actuator; it is not surprising that the translational stiffness increases when $k_{act}$ increases. The term $\frac{L_B}{2S_BE}$ shows the influence of virtual joint $8$ (parallelogram bars tension/compression): $K_b$ increases when $S_B$ or $E$ increase, and decreases when $L_B$ increases, which is in accordance with intuition;
  \item The term $\frac{4L_f^3\sin^2\lambda}{Eh_f^3b_f}$ shows the influence of the {\it foot related} virtual joints (tension/compression and bending): when $\lambda$ increases, with $\lambda \in [ 0 \; \pi/2 ]$ \footnote{If $\lambda \geq \pi/2$ the foot does not anymore ``move away'' the parallelogram from the prismatic actuated joint, which is one of its main functions (i.e. avoiding collisions between the actuator and the parallelogram); furthermore we must have $\lambda \geq 0$ to avoid interference between the foot and the actuated prismatic joint.}, then $\sin^2\lambda$ increases and consequently $K_b$ decreases. According to intuition, increasing $L_f$ decreases $K_b$, while increasing $h_f$ or $b_f$ increases $K_b$.
\end{itemize}
%%%%%%%%%%%%%%%%%%%%%%%%%%%%%%%%%%%%%%%%%%
\subsection{Quantitative analysis of $K_a$ and $K_b$}
%%%%%%%%%%%%%%%%%%%%%%%%%%%%%%%%%%%%%%%%%%
As we have seen, the qualitative analysis of $K_a$ and $K_b$ provides interesting information on the influence of the geometrical parameters on the rotational and translational stiffnesses. Quantitative information about the parameters' influence on the Orthoglide's stiffness can also be obtained from the symbolic expressions by studying the consequences of a -{~}100/+200\% variation of the parameters on $K_a$ and $K_b$. A variation of -100\% corresponds to a zero parameter, while +200\% corresponds to an extreme increase. Such a wide range of variation gives a global picture of the parameter's influence. The initial values of the parameters used for the computation are given in Tab.~\ref{parameters} and correspond to the dimensions of the prototype of the Orthoglide developed at IRCCyN. Parameters $k_{act}$ and $E$ are considered constant because our analysis is restricted to geometrical parameters only. We choose $E=7.10^4$ Nmm$^{-2}$ (aluminum) and $k_{act}=10^5$ Nmm$^{-1}$. The stiffness of the actuated prismatic joint depends on many parameters (mechanical components, electrical motor power, control). The chosen value is a commonly used one, however it is still much stiffer than the virtual joints, which is in accordance with our assumptions.

In order to clearly show the relative influence of each parameter, we are going to superimpose several curves on a same chart. Each curve represents a ratio $\frac{K_a(t)}{K_{a_{initial}}}$ (resp. $\frac{K_b(t)}{K_{b_{initial}}}$), in which $t$ is the percentage of variation of one of the parameters ($L_f$, $b_f$, $h_f$, $\lambda$, $L_B$, $S_B$ or d), while the other parameters remain at their initial value, and $K_{a_{initial}}$ (resp. $K_{b_{initial}}$) is the initial value of the torsional (resp. translational) stiffness when the parameters are at their initial value. Obviously, all $\frac{K_a(t)}{K_{a_{initial}}}$ (resp. $\frac{K_b(t)}{K_{b_{initial}}}$) curves cross when $t=0\%$.

\begin{figure}[ht!]
  \begin{center}
    \begin{tabular}{c c}
       \begin{minipage}[t]{70 mm}
          \begin{center}
            {\scalebox{.25}
            {\includegraphics{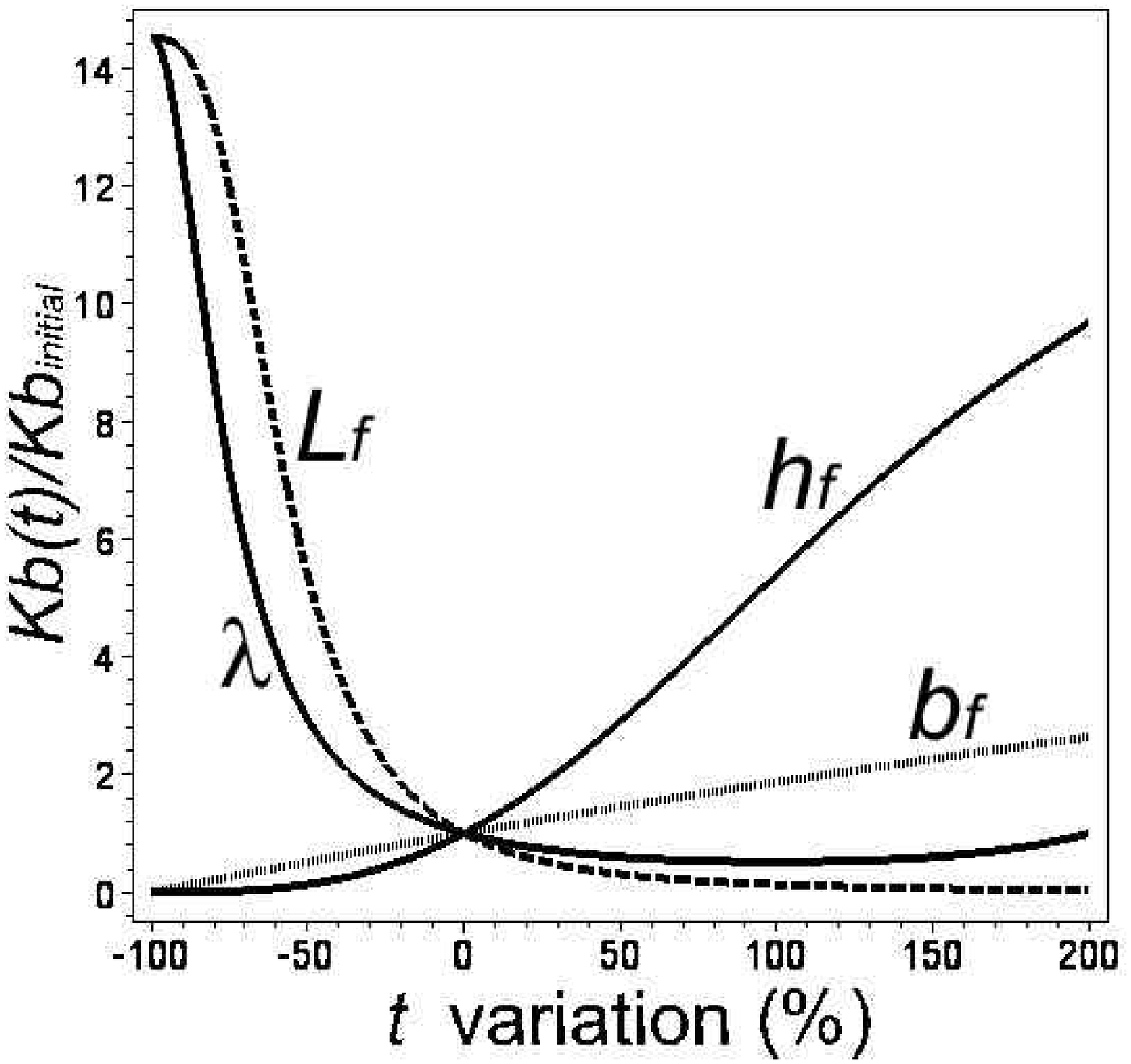}}}
          \end{center}
       \end{minipage} &
       \begin{minipage}[t]{70 mm}
          \begin{center}
           {\scalebox{.25}
           {\includegraphics{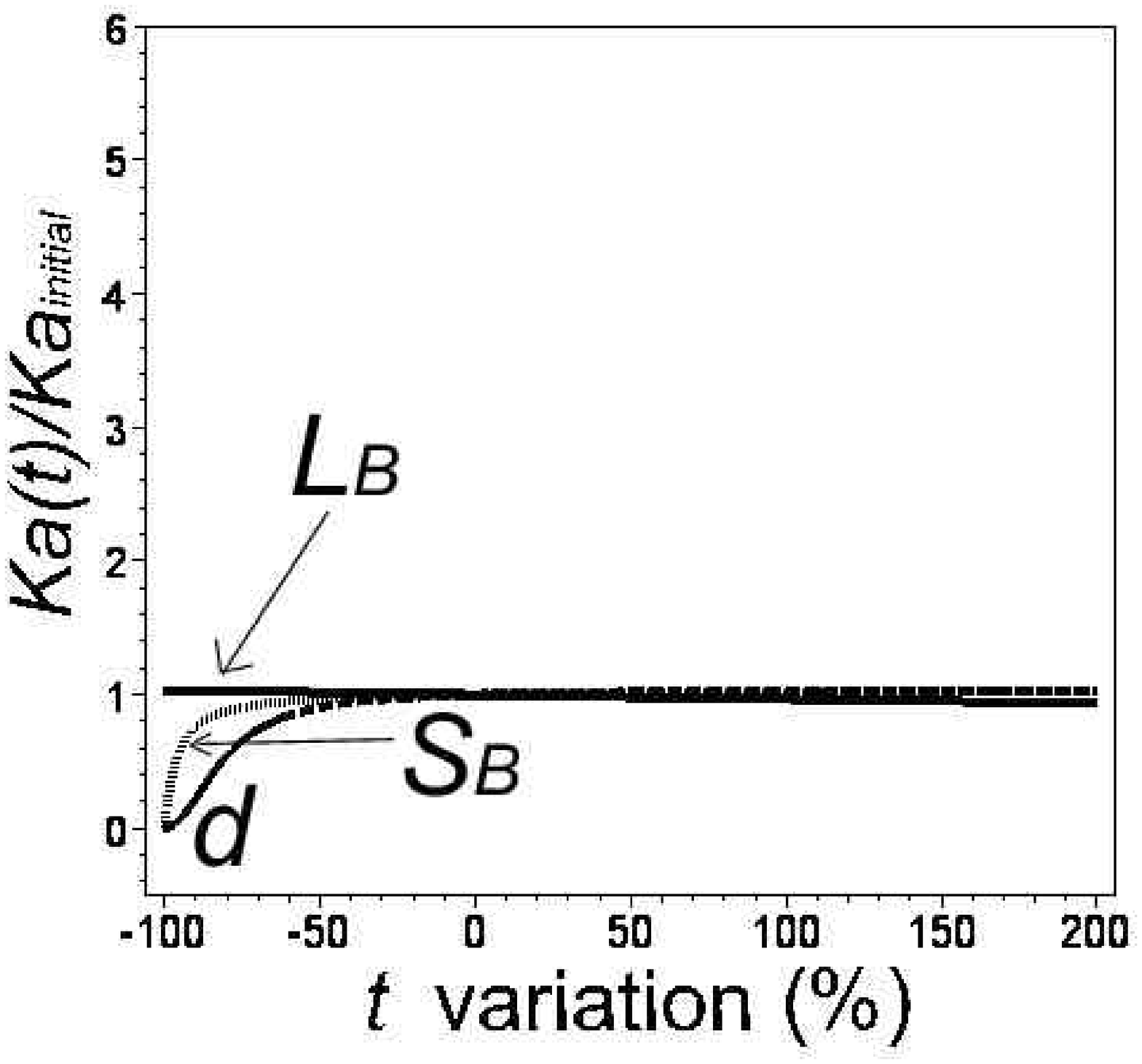}}}
          \end{center}
       \end{minipage} \\
       \begin{minipage}[t]{70 mm}
          \begin{center}
            {\small $K_a(t)/K_{a_{initial}}$: most influent parameters}
          \end{center}
       \end{minipage} &
       \begin{minipage}[t]{70 mm}
          \begin{center}
            {\small $K_a(t)/K_{a_{initial}}$: least influent parameters}
          \end{center}
       \end{minipage} 
        \end{tabular}
    \caption{Influence of the parameters on $K_a$} \label{Ka_isotropie}
\end{center}
\end{figure}

For example  let us replace each parameter in the symbolic expression of $K_a$ by its initial value except $L_f$. A one variable analytical expression $K_a(L_f)$ is then obtained:
 $$
  K_a(L_f)=\frac{0.56 \times 10^9}{L_f}
 $$
In this expression, let us replace $L_f$ by $L_{f_{initial}}(1+t)$. A new expression $K_a(t)$ is obtained:
 $$
  K_a(t)=\frac{0.56 \times 10^9}{150(1+t)}
 $$
where $t$ represents the percentage of variation of $L_f$. $K_a(t=0)$ gives the value for $K_{a_{initial}}$. We assume that $t$ varies from $-100\%$ to $+200\%$ as explained above. All $K_a(t)/K_{a_{initial}}$ curves obtained for all parameters are superimposed on a same chart so as to compare the parameters relative influence. %%%%%%%%%%%%%%%%%%%%%%%%%%%%%%%%%%%%%%%%%%
\begin{itemize}
  \item Quantitative analysis of $K_a$
\end{itemize}
%%%%%%%%%%%%%%%%%%%%%%%%%%%%%%%%%%%%%%%%%%
Figure~\ref{Ka_isotropie} shows the influence of the parameters on $K_a$. $L_B$, $d$ and $S_B$ have little influence compared to $L_f$, $h_f$, $b_f$ and $\lambda$.

\begin{figure}[ht!]
  \begin{center}
          \scalebox{.45}
          {\includegraphics{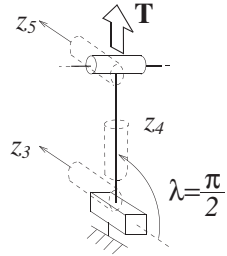}}
          \caption{Only virtual joints $4$ of the foot is affected by {\bf T} when  $\lambda=\pi/2$}
          \label{jambe_flex_90}
  \end{center}
\end{figure}

$K_a(\lambda)$ is a maximum (52\% increase) when $\lambda$ increases by 100\%, i.e. when $\lambda=\pi/2$. This result can also be obtained through observation of the symbolic expression of $K_a$: indeed, the initial values of $h_f$ and $b_f$ ($h_f=26$, $b_f=16$) make $(45h_f^2-33b_f^2)$ positive. Therefore, the denominator of
$K_a$ will be a minimum when $\lambda=\pi/2$. Moreover, when $\lambda=\pi/2$, the torque {\bf T} that is transmitted by the leg no longer has a component along the axis of virtual joints $3$ and $5$ of the foot (Fig. \ref{jambe_flex_90}). This is a physical explanation for $K_a(\lambda)$ being maximum when $\lambda=\pi/2$.

Furthermore, $K_a$ increases more with $b_f$ than with $h_f$ for a same variation. Consequently, for a given foot weight increase, the torsional stiffness benefits more from an increase of $b_f$ than from an increase of $h_f$. From a designer's point of view, this is valuable information. If the foot length $L_f$ increases, $K_a$ decreases since in this case the {\it foot and torque related} stiffnesses $k_3$, $k_4$, $k_6$ decrease. Conversely if $L_f$ decreases then $K_a$ increases tremendously.

\begin{figure}[ht!]
  \begin{center}
  \begin{tabular}{c c}
       \begin{minipage}[t]{60 mm}
         \begin{center}
        {\scalebox{.25}
        {\includegraphics{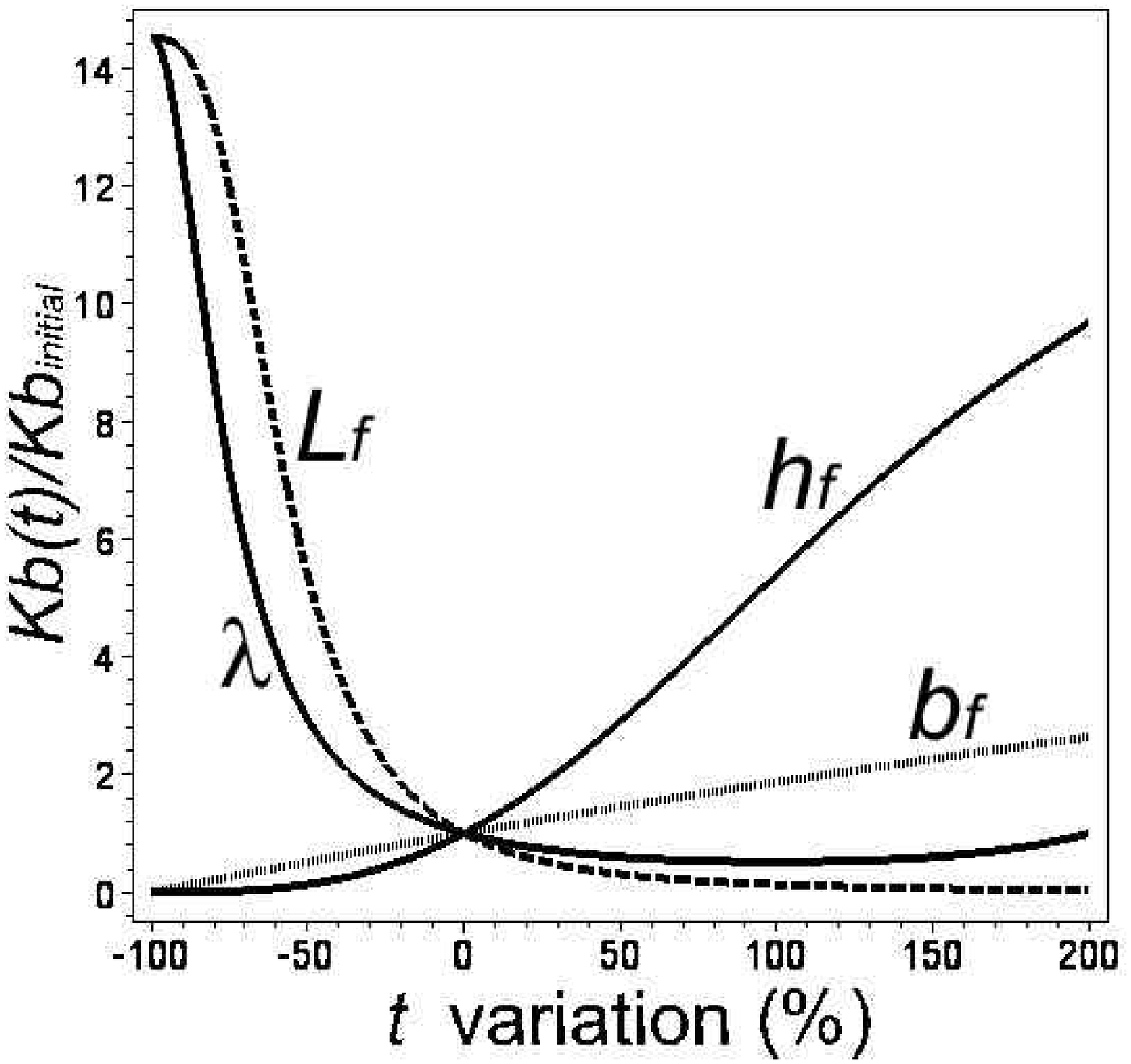}}}
         \end{center}
       \end{minipage} &
       \begin{minipage}[t]{60 mm} 
         \begin{center}
        {\scalebox{.25}
        {\includegraphics{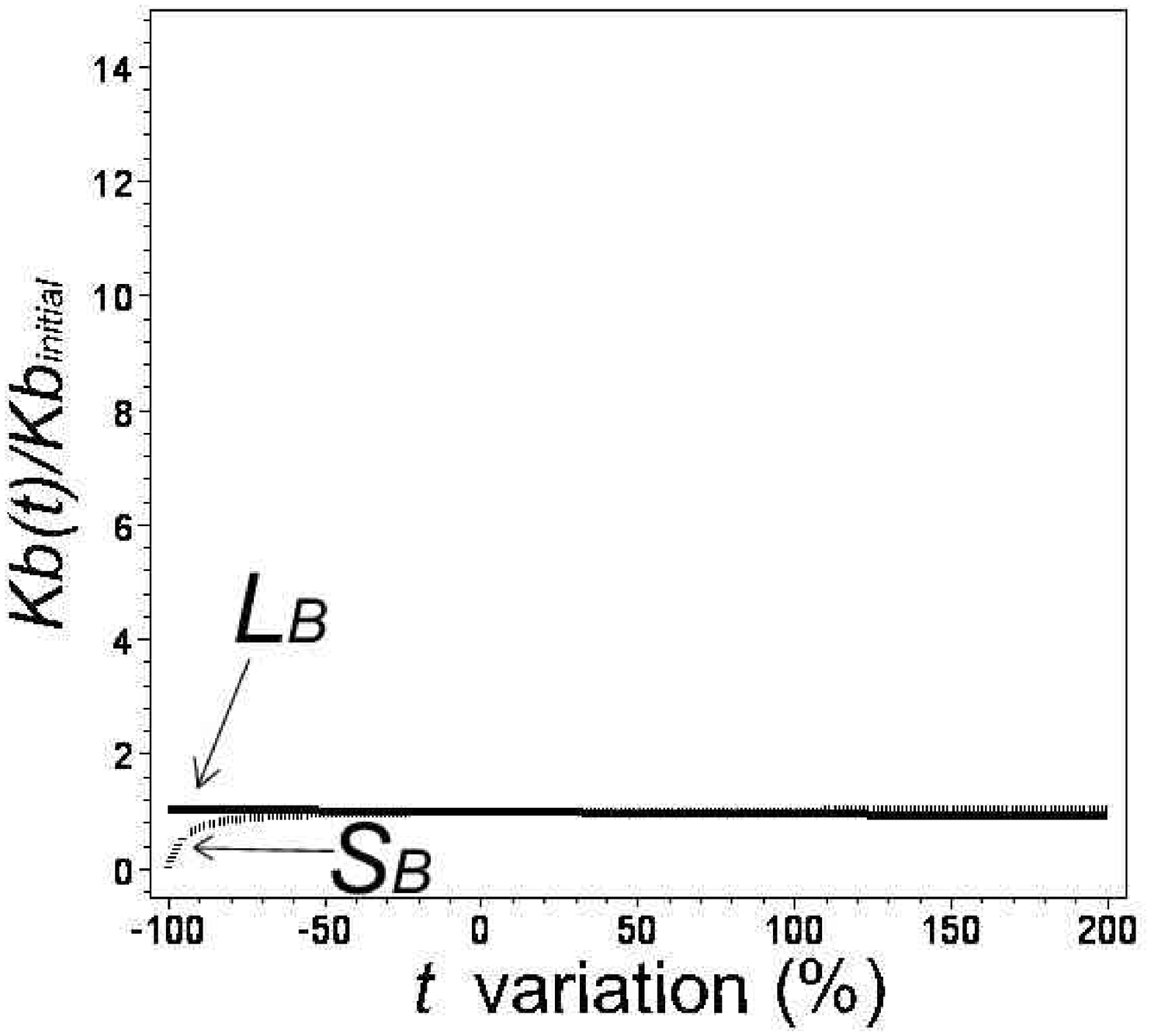}}}
         \end{center}
       \end{minipage} \\
       \begin{minipage}[t]{70 mm}
         \begin{center}
        {\small $K_b(t)/K_{b_{initial}}$: most influent parameters}
         \end{center}
       \end{minipage} &
       \begin{minipage}[t]{70 mm} 
         \begin{center}
        {\small $K_b(t)/K_{b_{initial}}$: least influent parameters}
         \end{center}
       \end{minipage} 
        \end{tabular}
\caption{Influence of the parameters on $K_b$} \label{Kb_isotropie}
\end{center}
\end{figure}

Finally, we observe that when $d$, $S_B$, $h_f$ or $b_f$ tend towards zero, then so does $K_a$. This can be deduced from the symbolic expression of $K_a$, but also tends to a physical interpretation: if the foot or the parallelogram bars tend to have a very small cross section, or if the parallelogram tends not to be able to support any torque (when $d$ tends towards zero), then the whole mechanism loses its torsional stiffness. Though $h_f$ and $\lambda$ play important roles in $K_a$, the two most important parameters are $L_f$ and $b_f$.
%%%%%%%%%%%%%%%%%%%%%%%%%%%%%%%%%%%%%%%%%%
\begin{itemize}
  \item Quantitative analysis of $K_b$
\end{itemize}
%%%%%%%%%%%%%%%%%%%%%%%%%%%%%%%%%%%%%%%%%%
Figure~\ref{Kb_isotropie} shows the influence of the geometrical parameters on $K_b$. We can observe that $L_B$ and $S_B$ have little influence compared to $L_f$, $h_f$, $b_f$ and $\lambda$. $K_b(\lambda)$ is a minimum (48\% decrease) for a $100$\% increase of $\lambda$, i.e. when $\lambda=\pi/2$. This conclusion can be reached through the observation of the symbolic expression of $K_b$: indeed, we can see that the denominator of $K_b$ will be a maximum when $\lambda=\pi/2$.

\begin{figure}[ht!]
  \begin{center}
           {\scalebox{.45}
           {\includegraphics{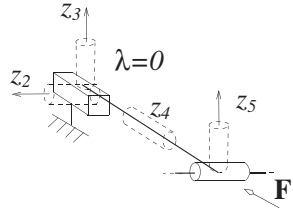}}}
           \caption{Only virtual joint $3$, $4$ and $5$ of the foot is affected by ${\bf F}$ when $\lambda=0$}
           \label{jambe_flex_00}
  \end{center}
\end{figure}

From the symbolic expression of $K_b$, one can also infer that if $\lambda$ decreases, then the denominator will decrease and consequently $K_b$ will increase. This was the opposite case for $K_a$. For a $100$\% decrease of $\lambda$, $K_b$ will be a maximum: 14.4 times its initial value. This has a physical interpretation: when $\lambda=0$, the virtual joint $2$ is no longer affected by the force {\it \bf F} that is transmitted by the leg (Fig. \ref{jambe_flex_00}).

$K_b$ is also a maximum (14.4 times its initial value) when $L_f=0$. However, the physical interpretation is not the same. When $L_f=0$, the stiffness of the virtual joints $2$, $3$, $4$ and $5$ tends toward $+\infty$ which makes them behave like infinitely stiff virtual joints, making the mechanism as a whole
much more stiffer. One can also observe that $K_b$ increases more with $h_f$ than
with $b_f$. This can be concluded from the symbolic expression of $K_b$.
Indeed, from Eq. \ref{stiffmat_iso_tors}, we have:
$$
  K_b(h_f,b_f)=\frac{1}{0.00002537698413+\frac{96.42857143}{h_f^3b_f}}
$$
Consequently, a 10\% increase of $h_f$ will make the denominator of $K_b$ decrease faster than a 10\% increase of $b_f$. Finally, if $S_B$, $h_f$ or $b_f$ tend towards zero, then $K_b$ also tends towards zero. This can be concluded from the symbolic expression of $K_b$, but it also corresponds to the physical phenomenon that was explained for $K_a$.

The most important parameters for $K_b$ are $\lambda$, $L_f$ and $h_f$. Parameters $\lambda$ and $L_f$ have a similar influence: when they decrease, $K_b$ increases, and conversely. Parameter $h_f$ has the opposite influence: when $h_f$ increases, $K_b$ increases, and conversely. The symbolic expressions of $K_b$ as univariate functions of these three parameters are of great help at a pre-design stage to analyze the translational stiffness.
%%%%%%%%%%%%%%%%%%%%%%%%%%%%%%%%%%%%%%%%%%
\subsection{Conclusions}
%%%%%%%%%%%%%%%%%%%%%%%%%%%%%%%%%%%%%%%%%%
The analysis of the symbolic expressions of $K_a$ and $K_b$ at the isotropic configuration allows us to plot the most influent parameters in this configuration, and the way their variation influences the mechanism's stiffness. A global analysis must be conducted in the whole workspace to determine the global influence of the parameters. This was achieved in \cite{Majou04a}, with the determination of a line along which the stiffness analysis results hold for the whole workspace. Such a procedure is required to simplify the global stiffness analysis. However, as mentioned above in the case of the Orthoglide, analyzing the stiffness at the isotropic configuration can give a good overview of the performances.

The use of simple symbolic expressions allows us to deduce helpful results in order to improve the Orthoglide's stiffness. However these modifications must be made while taking into account the technological constraints (collisions, interferences) of the prototype initial architecture. For example, if one sets $\lambda$ to zero in order to increase $K_b$, then the offset between kinematic joints $L_{6}$ and $L_{1}$ disappears. However, this offset aims at preventing the parallelogram from colliding with the prismatic actuated joint. Therefore it is not possible to set $L_f$ or $\lambda$ to zero. It is better, either to only lower them and check how much the reachable workspace is then reduced, or to increase $h_f$, or both. Conversely, if one wants to increase the collision-free workspace by increasing $L_f$ while keeping $K_b$ constant, studying the simultaneous influence on the stiffness of $L_f$ and $h_f$ or $L_f$ and $b_f$ can then prove useful. One problem will be the foot weight increase that will require more powerful actuators to keep the dynamic performances at a similar level. We consider the issue of simultaneous variation of two parameters in the following section.
%%%%%%%%%%%%%%%%%%%%%%%%%%%%%%%%%%%%%%%%%%
\section{Influence of the simultaneous variation of two parameters}
\begin{figure}[ht!]
\begin{centering}
  \begin{tabular}{c c}
       \begin{minipage}[t]{60 mm}
         \begin{center}
         {\scalebox{.28}
         {\includegraphics{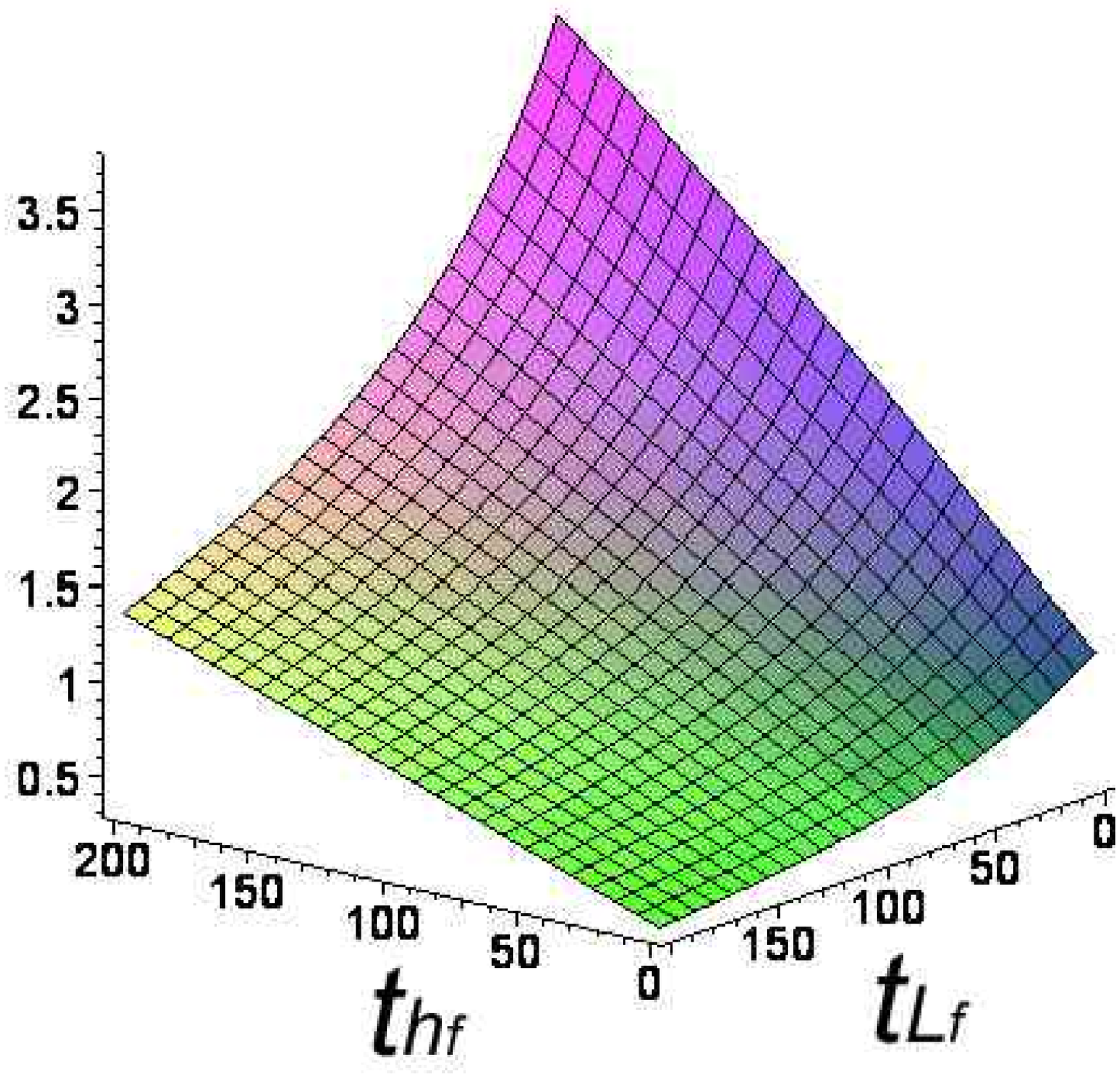}}}
         \end{center}
       \end{minipage} &
       \begin{minipage}[t]{60 mm}
         \begin{center}
         {\scalebox{.25}
         {\includegraphics{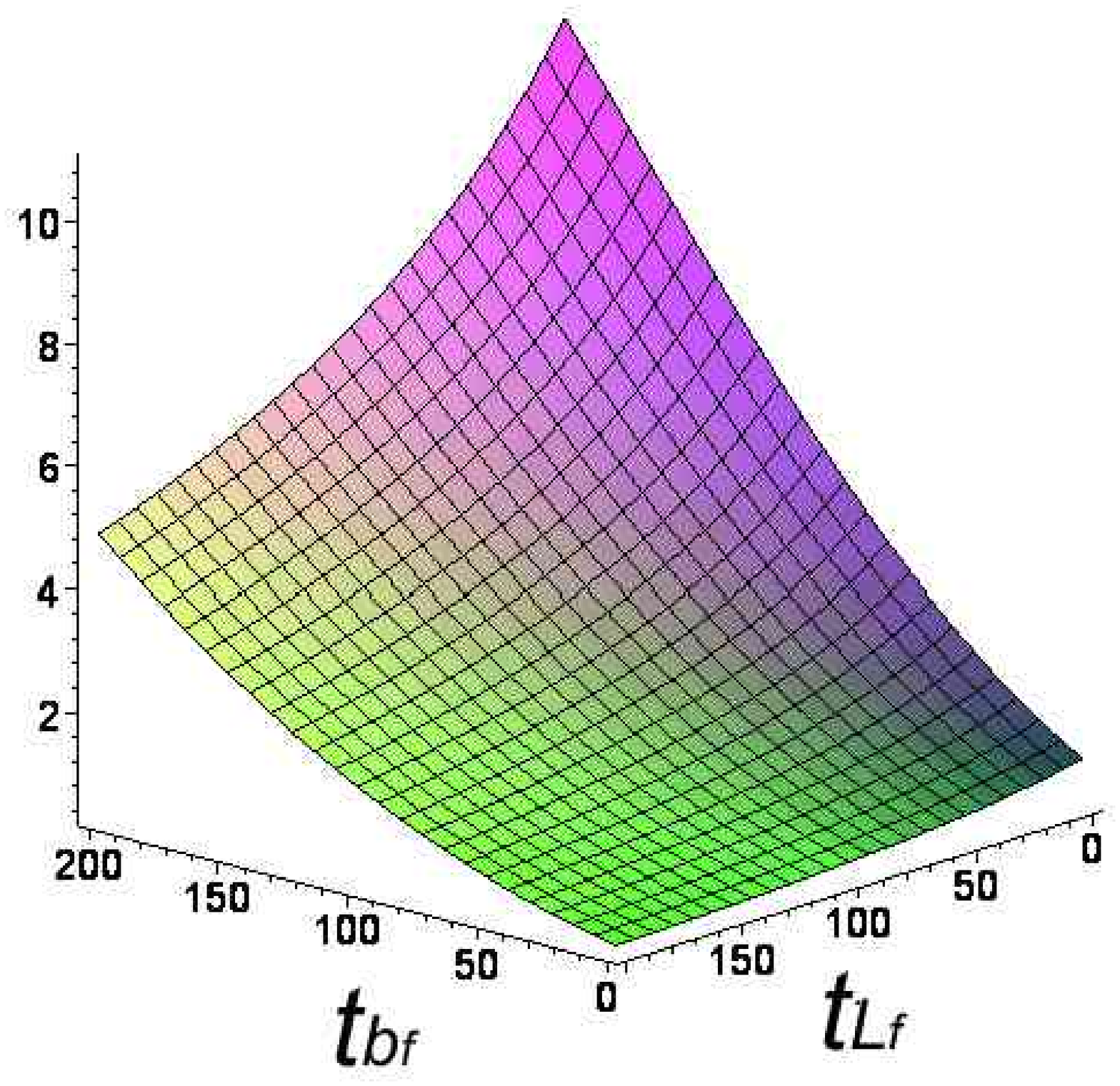}}}
         \end{center}
       \end{minipage} \\
       \begin{minipage}[t]{60 mm}
         \begin{center}
         {\small $K_a(t_{h_f},t_{L_f})/K_{a_{initial}}$}
         \end{center}
       \end{minipage} &
       \begin{minipage}[t]{60 mm}
         \begin{center}
         {\small $K_a(t_{b_f},t_{L_f})/K_{a_{initial}}$}
         \end{center}
       \end{minipage}
    \end{tabular}
  \caption{$K_a/K_{a_{initial}}$ as a function of $h_f$, $b_f$ and $L_f$}
  \label{KaKb_simultane}
  \end{centering}
\end{figure}

\begin{figure}[ht!]
\begin{centering}
  \begin{tabular}{c c}
       \begin{minipage}[t]{60 mm}
         \begin{center}
            {\scalebox{.25}
            {\includegraphics{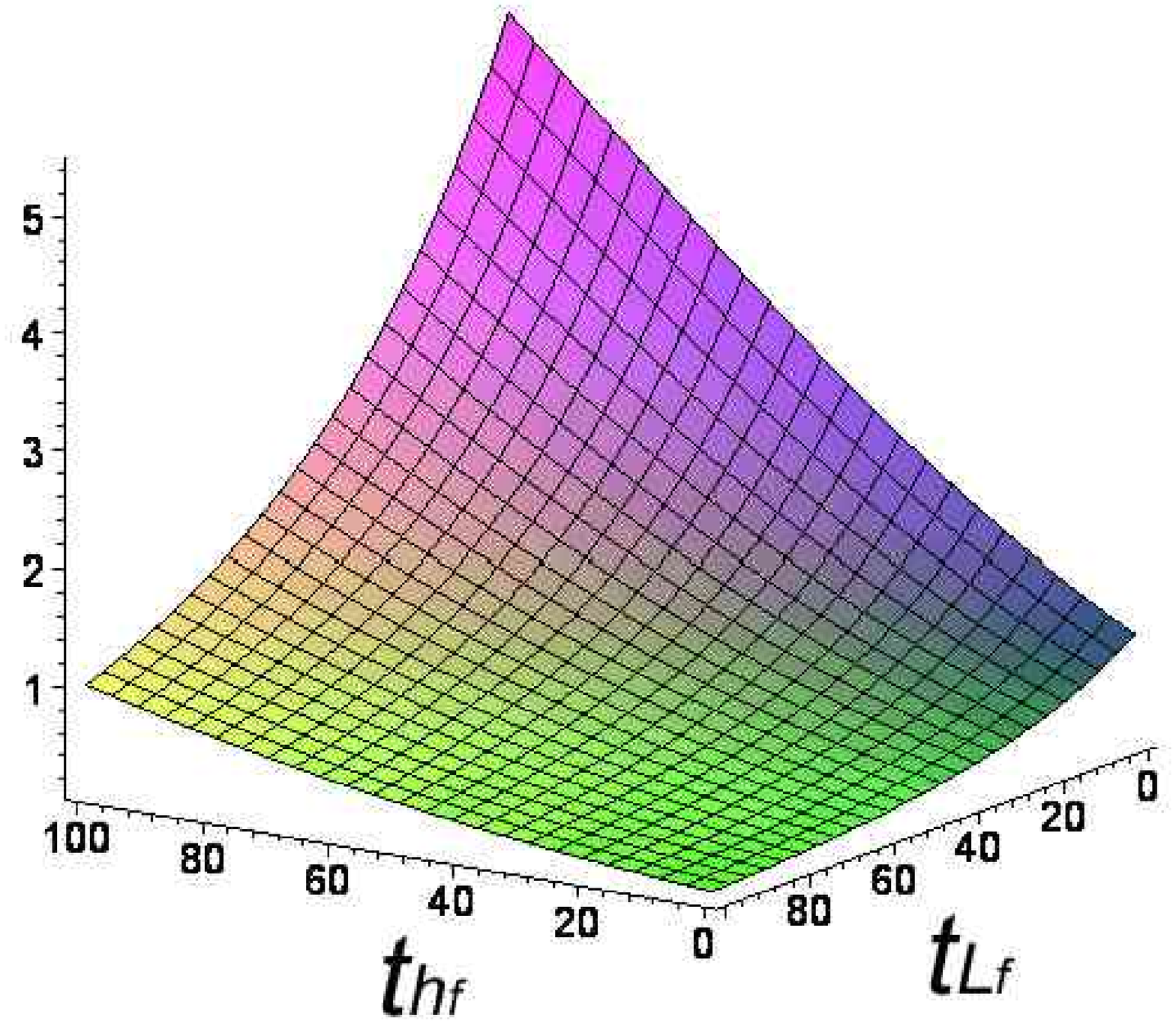}}}
         \end{center}
       \end{minipage} &
       \begin{minipage}[t]{60 mm}
         \begin{center}
          {\scalebox{.25}
          {\includegraphics{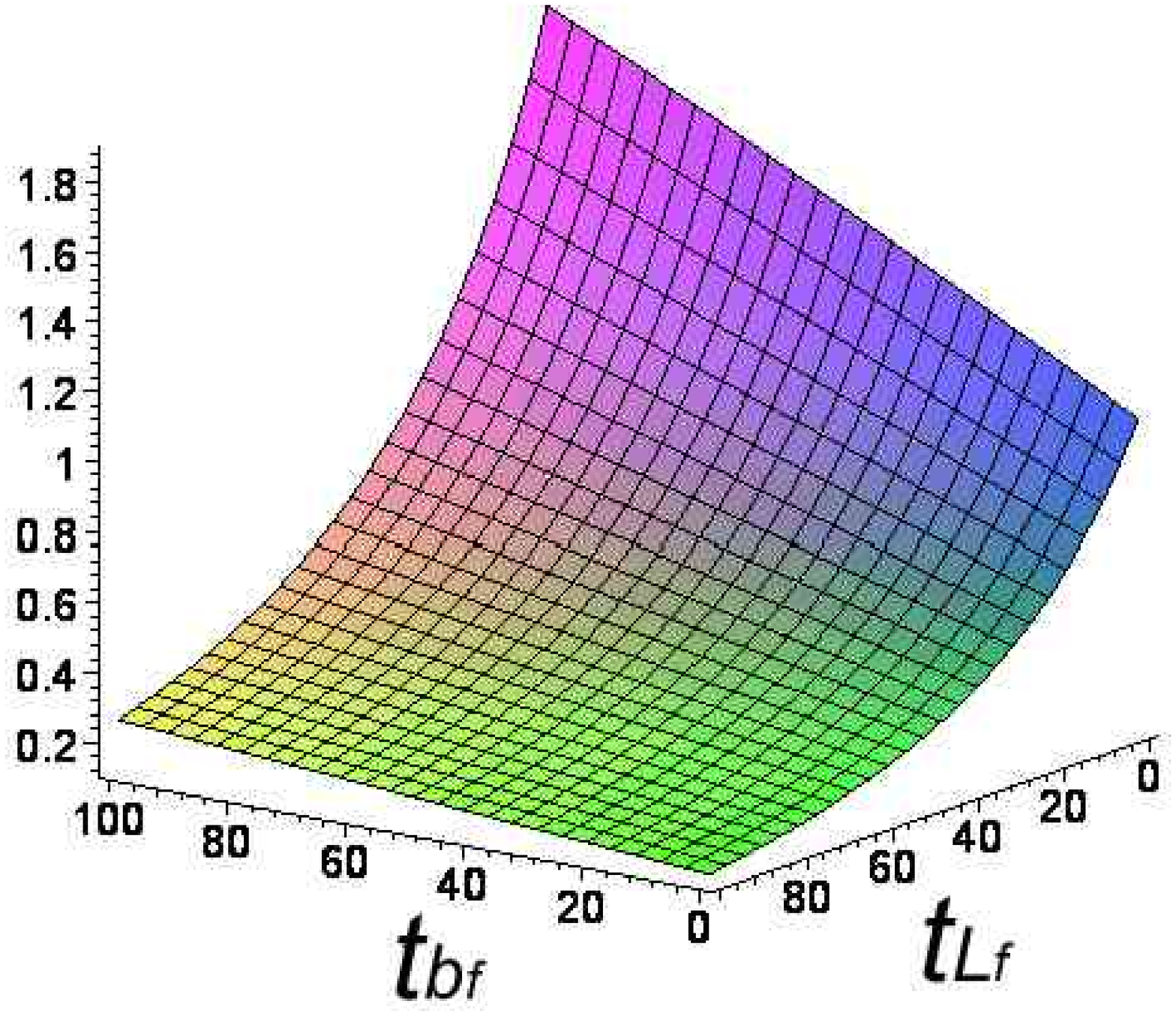}}}
         \end{center}
       \end{minipage} \\
       \begin{minipage}[t]{60 mm}
         \begin{center}
           {\small $K_b(t_{h_f},t_{L_f})/K_{b_{initial}}$}
         \end{center}
       \end{minipage} &
       \begin{minipage}[t]{60 mm}
         \begin{center}
           {\small $K_b(t_{b_f},t_{L_f})/K_{b_{initial}}$}
         \end{center}
       \end{minipage}
   \end{tabular}
   \caption{$K_b/K_{b_{initial}}$ as a function of $h_f$, $b_f$ and $L_f$}
   \label{KaKb_simultane_1}
  \end{centering}
\end{figure}

%%%%%%%%%%%%%%%%%%%%%%%%%%%%%%%%%%%%%%%%%%
In this section, we study the influence on $K_a$ and $K_b$ of the simultaneous variation of two parameters $L_f$ and $h_f$, or $L_f$ and $b_f$, at the isotropic configuration. Analytical expressions of $K_a$ and $K_b$ as functions of two variables are deduced from Eq.~\ref{stiffmat_iso_tors}. Figures~\ref{KaKb_simultane} and \ref{KaKb_simultane_1} show plots of $K_a/K_{a_{initial}}$ and $K_b/K_{b_{initial}}$ when $t_{h_f}$ (resp. $t_{b_f}$) --- which is the relative variation of $h_f$ (resp. $b_f$) --- and $t_{L_f}$ --- which is the relative variation of $L_f$ --- increase from 0 to $100$\% or $200$\% when relevant.

Figure~\ref{KaKb_simultane} shows that increasing $h_f$ or $b_f$ allows us to compensate for the decrease of $K_a$ occurring when $L_f$ increases. For example if $L_f$ increases by $50\%$, $h_f$ must increase by $34\%$ or $b_f$ must increase by $16\%$ for $K_a$ to remain at its initial value $K_{a_{initial}}$. Regarding the dynamic performances (i.e. the foot weight increase), it will be more interesting to increase $b_f$ by $16\%$.

On Fig. \ref{KaKb_simultane_1}, one can also observe that increasing $h_f$ or $b_f$ allows us to easily compensate for the decrease of $K_b$ occurring when $L_f$ increases. If $L_f$ increases by $50\%$, $h_f$ must increase by $48\%$ or $b_f$ must increase by $245\%$ for $K_b$ to remain at its initial value $K_{b_{initial}}$. Therefore, it seems more judicious to increase $h_f$ rather than $b_f$ in order to compensate for the stiffness loss due to the increase of $L_f$, because the foot weight increase is lower. This is a multi-criteria multi-parameters ($L_f$, $h_f$, $b_f$) optimization problem: increasing the collision-free workspace while keeping the same stiffness, with a minimum foot weight increase. Integrating the symbolic expressions of the stiffness in multicriteria optimization loops could be an interesting extension of our work.
%%%%%%%%%%%%%%%%%%%%%%%%%%%%%%%%%%%%%%%%%%
\section{Analysis of the tool displacements induced by external forces}
%%%%%%%%%%%%%%%%%%%%%%%%%%%%%%%%%%%%%%%%%%
Another interesting use of the symbolic expressions of $\kappa_{ij}$ is to observe the tool compliant displacements when simulated cutting forces are applied on the tool. By multiplying these forces with the compliance matrix and analyzing the evolution of the compliant displacements obtained, as a function of the Cartesian coordinates, the stiffest zones of the mechanism's workspace can be determined. Thus, the global stiffness behavior is taken into account. As the simulated cutting forces correspond to a particular manufacturing operation, the stiffest zone will be specific to the application.The equations with which the stiffness matrix is computed are built using the principle of virtual work. Simulated cutting forces will then correspond to quasi-static conditions, which may not be realistic in some cases. In this section, a simple groove milling operation is simulated, which can be considered as a quasi-static operation.

The symbolic derivation of the stiffness matrix $\bf K$ using the method described above was achieved with Maple software on a 1 GHz, 256MB RAM PC. The computation of $\bf K$ did not end within one day, which means that the components of matrix $\bf K$, i.e. the $K_{ij}$, are too large to be manipulated within a Maple worksheet. However, computing the components of the compliance matrix $\bb \kappa$, i.e. the $\kappa_{ij}$, took 12 hours only. This resulted in symbolic expressions that remained relatively easy to manipulate within a Maple worksheet. Therefore we choose to analyze the Orthoglide's stiffness through the analysis of the symbolic expressions of the $\kappa_{ij}$: the main idea is that when the $\kappa_{ij}$ increase, then the Orthoglide's stiffness decreases.
%%%%%%%%%%%%%%%%%%%%%%%%%%%%%%%%%%%%%%%%%
\subsection{Compliant displacements}
\begin{figure}[hb!]
    \begin{center}
          \scalebox{.45}{\includegraphics{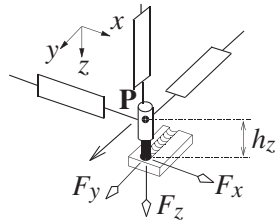}}
           \caption{Component forces of groove milling operation}
           \label{groove_milling}
    \end{center}
\end{figure}

%%%%%%%%%%%%%%%%%%%%%%%%%%%%%%%%%%%%%%%%%%
Vector $\bf w$ is the static wrench of the cutting forces applied on the tool during the groove milling operation along the $y$ axis. We have:
$$
  {\bf w}=\left\{ \begin{array}{c}
             {\bf T}\\
             {\bf F}
                   \end{array}
          \right\}
  \quad {\rm with} \quad
  {\bf F} = [F_x~F_y~F_z]^T
  \quad {\rm and} \quad
  {\bf T} = [-F_yh_z~F_xh_z~0]^T \quad  (Fig. \ref{groove_milling})
$$
The compliant displacements of the mobile platform are computed as follows:
%%% CD
$$
%  {\bf t}  = {\bb \kappa} {\bf w}
  {\bf d}  = {\bb \kappa} {\bf w}
  \quad 
  {\rm with}  
  \quad
%     {\bf t} =\left\{ \begin{array}{c}
     {\bf d} =\left\{ \begin{array}{c}
                       {\bb \Omega}\\
                       \bf{V}
                     \end{array}
             \right\},
  \quad
  {\bb \Omega} = [{\omega}_x~{\omega}_y~{\omega}_z]^T
  \quad {\rm and}
  \quad
   {\bf V} = [v_x~v_y~v_z]^T
$$
The compliant displacements at the tool tip are then:
$$
%{\bf t_{tool}} =\left\{ \begin{array}{c}
{\bf d_{tool}} =\left\{ \begin{array}{c}
             {\bb \Omega}\\
             \bf{V}+{\bb \Omega}\times {\it h_z}{\bf z}
                   \end{array}
          \right\}
$$
%%%%%%%%%%%%%%%%%%%%%%%%%%%%%%%%%%%%%%%%%%
\subsection{Determination of the stiffest working zones for a given task}
%\vspace{-5mm}
%%%%%%%%%%%%%%%%%%%%%%%%%%%%%%%%%%%%%%%%%%
With the symbolic expressions of the tool displacement, one can evaluate the tracking error along the groove path. Using the symbolic expression of the tracking error, a stiffness favorable working zone, i.e. a working zone in which the tracking error is low, can be determined. To simulate cutting forces during the groove milling operation, a High Speed Machining (HSM) simulation software is used \cite{Dugas02}. Depending on the manufacturing conditions, this software provides the average cutting forces. The manufacturing conditions chosen for the groove milling are:
  \begin{itemize}
    \item Spindle rate is N=20,000tr.min$^{-1}$; 
    \item Feed rate V$_f$=40m.min$^{-1}$;   
    \item Cutting thickness is 5.10$^{-3}$mm; 
    \item The tool is a ball head of $\Phi=10$ mm diameter with 2 steel blocks; 
    \item Manufactured material is a common steel alloy with chromium and molybdenum.
  \end{itemize}
The simulated cutting forces correspond to a HSM context, which is what PKM are $F_x=215 N$, $F_y=-10 N$, $F_z=-25 N$. The above data allows us to simulate the tool compliant displacement along a groove path along the $y$ axis (see Fig. \ref{groove_milling}). $h_z$=100 mm corresponds to the tool mounted on the prototype of the Orthoglide. The tracking error is the projection of the tool compliant displacement in the plane that is perpendicular to the path. We specify one groove path with its coordinates $(x_t,z_t)$, and one point ${\bf P}$ with $(x_t,y_P,z_t)$ coordinates located along this trajectory. The tracking error at point ${\bf P}$ is defined as $\delta_P = \sqrt{v_x^2+v_z^2}$.
%\begin{figure}[ht!]
%  \begin{center}
%       {\scalebox{.7}
%       {\includegraphics{Fig13.eps}}
%       \caption{Computation of the tracking error}
%       \label{erreur_poursuite}} 
%  \end{center}
%\end{figure}

\begin{figure}[ht!]
  \begin{center}
    \begin{tabular}{c c}
       \begin{minipage}[t]{70 mm}
       {\scalebox{.35}
       {\includegraphics{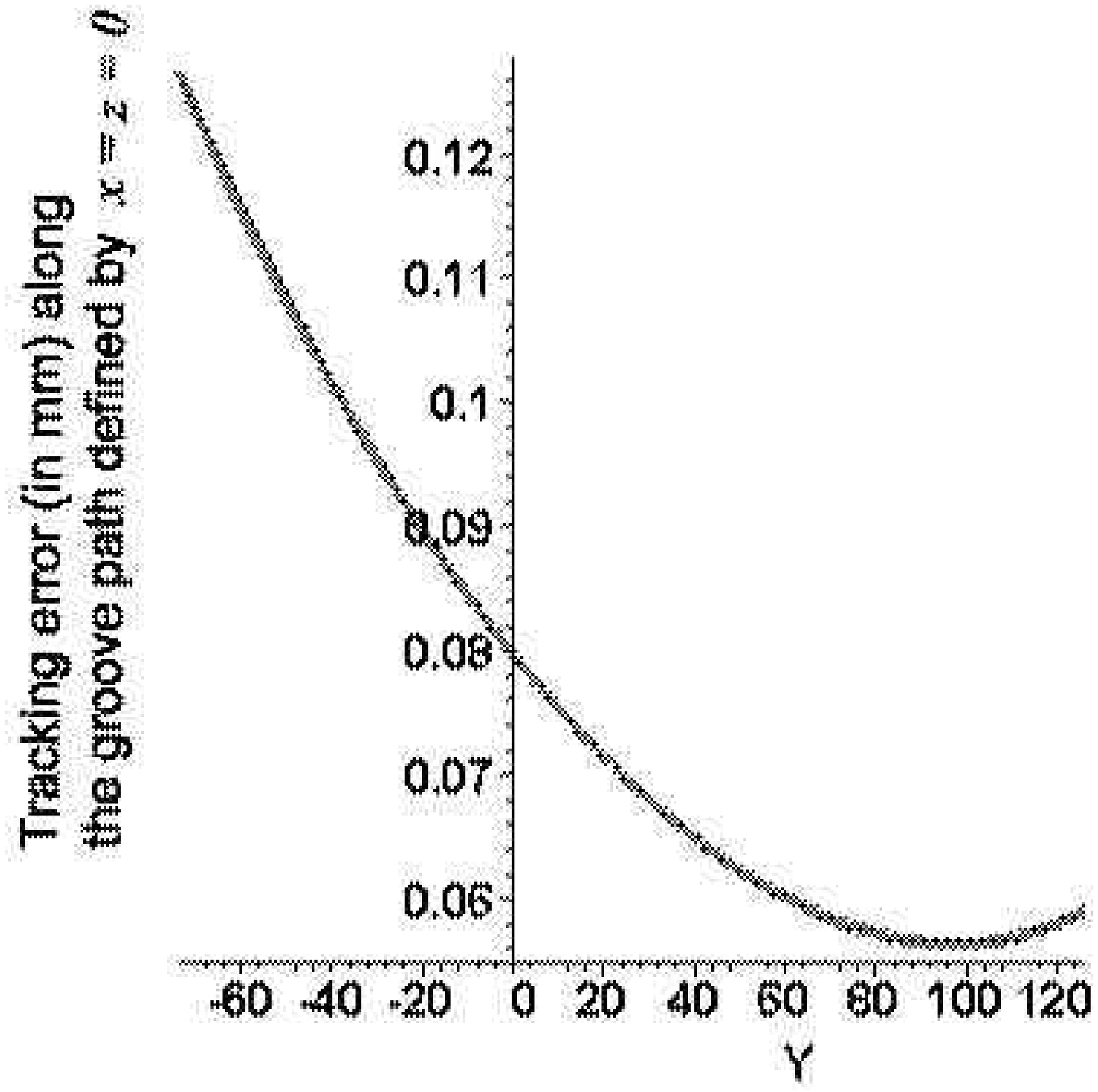}}
       \caption{Tracking error along the groove path defined by $x=z=0$}
       \label{erreur_poursuite_XZegale0}}
       \end{minipage} &
       \begin{minipage}[t]{70 mm}
         {\scalebox{.4}
         {\includegraphics{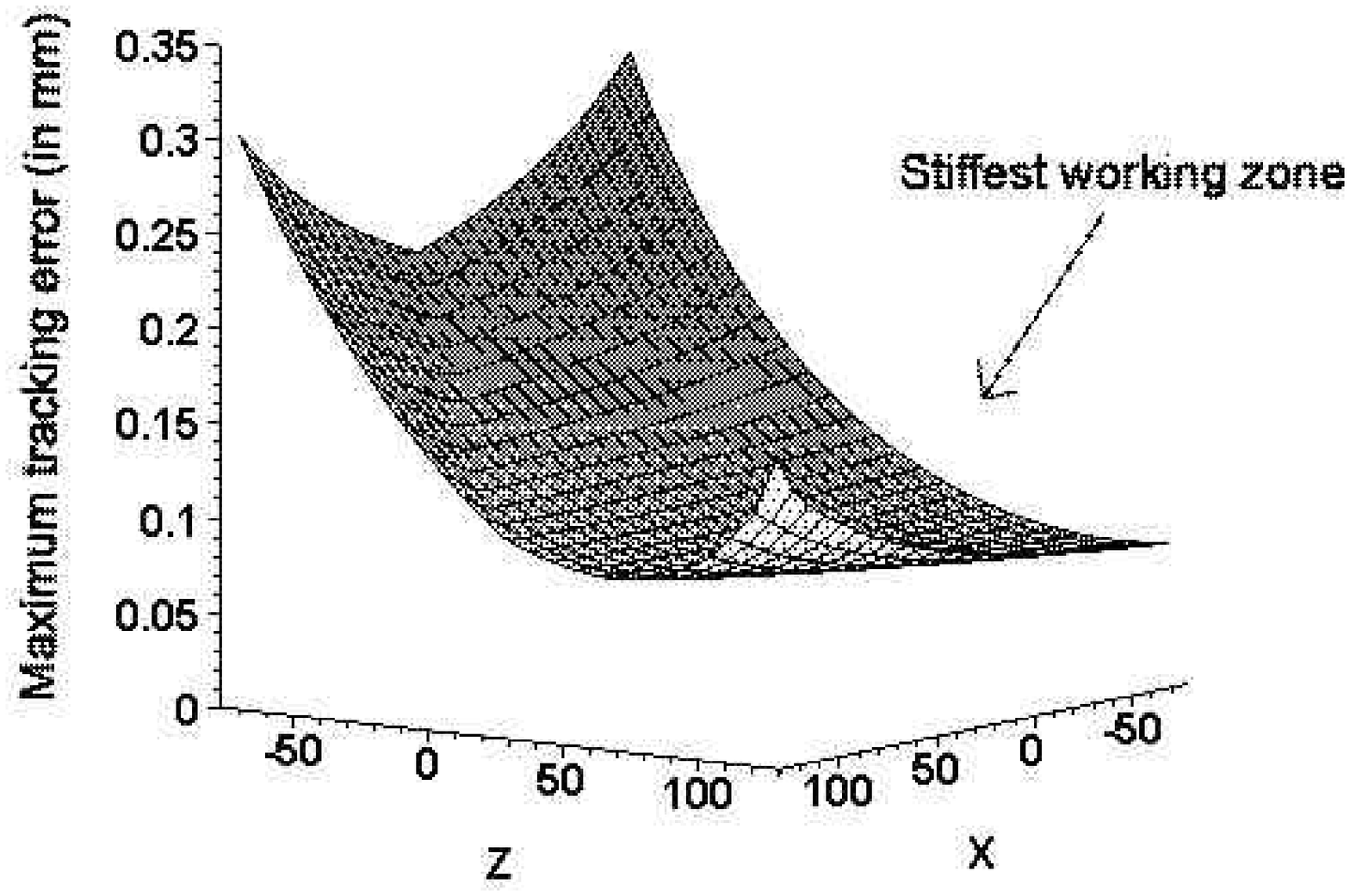}}
           \caption{Maximum tracking error along $y$ axis groove paths}}
           \label{erreur_poursuite_maximale}
       \end{minipage}
    \end{tabular}
  \end{center}
\end{figure}

The paths are defined in a cube centered at the intersection of the prismatic joints, $x_t$ and $z_t$ vary within the interval $[-73.65;126.35]$. We noticed that the maximum tracking errors were always located at one of the path ends. Figure~\ref{erreur_poursuite_XZegale0} shows the tracking error along a groove defined with the coordinates $(x_t,z_t)=(0,0)$. We can see that the maximum error occurs when $y=-73.65$, i.e. at one of the path ends. Depending on the coordinates $(x_t,z_t)$, the maximal tracking error is located at $y=-73.65$ or at $y=126.35$. Figure~\ref{erreur_poursuite_maximale} shows the maximum tracking error for each groove path defined by its coordinates $(x_t,z_t)$. The results clearly show a zone in which the maximum tracking error is low. In this working zone, $x$ varies within the interval [-73.65;0] and $z$ varies within [50;126.35]. It is difficult to find a physical explanation for this result. It depends on the cutting forces applied, their magnitude and direction, and on each virtual joint reaction to the wrench transmitted by the leg, which depends on the Cartesian coordinates. The information obtained, i.e., the lowest tracking error working zone, is, however, of great interest for the end-user in order to place manufacturing paths in the workspace achieving the lowest tracking error due to structural compliance.

Another use of the symbolic expressions of the compliant displacements would be the optimization of the geometric parameters to minimize the tracking error for specific cutting forces. This would mean optimizing a PKM design for a specific task. Our opinion is that it is better to look for global stiffness improvement as we did in the previous section. This way, optimization brings stiffness improvement to all potential manufacturing tasks. However, given a PKM design, it is very interesting to determine the stiffest working zones for specific tasks, as we did in this section.
%%%%%%%%%%%%%%%%%%%%%%%%%%%%%%%%%%%%%%%%%%
%\vspace{-5mm}
\section{Comparison with a Finite Element Stiffness Model}
%\vspace{-5mm}
%%%%%%%%%%%%%%%%%%%%%%%%%%%%%%%%%%%%%%%%%%
By comparing our stiffness model with a Finite Element Model (FEM) of the Orthoglide prototype, we will now show that our rigid link model is reasonably realistic \cite{Majou04a}. A FEM was implemented in LARAMA (LAboratoire de Recherches en Automatique et M\'ecanique Avanc\'ee, Clermont-Ferrand, France) as part of a collaboration within project ROBEA, a research program sponsored by CNRS (Centre National de la Recherche Scientifique). Due to space limitations, the modeling assumptions of the FEM are not detailed here. The FEM allows to calculate the variation range of diagonal elements ${\bf Kt}_{1,1}$, ${\bf Kt}_{2,2}$, ${\bf Kt}_{3,3}$ of translational stiffness  matrix ${\bf Kt}$, based on a CAD model of the Orthoglide implemented in the finite element software ANSYS \cite{Abiven:2002}.
The results obtained are presented in Tab. \ref{comp_MEF} (deterministic approximations and variation ranges) at the isotropic configuration. Our objective is to compare these results to those obtained with our Rigid Link Compliant Model (RLCM). Stiffnesses are expressed in N.mm$^{-1}$.
\begin{table}[ht!]
  \begin{centering}
  \begin{tabular}{|| c || c | c | c | c | c | c ||}
  \hline
  {~} & \multicolumn{2}{c|}{${\bf Kt}_{1,1}$} &
  \multicolumn{2}{c|}{${\bf Kt}_{2,2}$} &
  \multicolumn{2}{c||}{${\bf Kt}_{3,3}$}\\
  \cline{2-7}
  {~} & {FEM} & {RLCM} & {FEM} & {RLCM} & {FEM} & {RLCM} \\ \hline
  Isotropic configuration & 3500 & 2715 & 3500/4000 & 2715 & 3500/4000 & 2715 \\
\hline
\end{tabular}
\caption{Comparison of the RLCM and the FEM} \label{comp_MEF}
\end{centering}
\end{table}
The numbers obtained from the FEM are comparable to those obtained from the RLCM. Even if deterministic values are not equal, this comparison shows that the RLCM of the Orthoglide is reliable enough for the purposes of pre-design. However, a more detailed FEM analysis and experimental results based on the Orthoglide prototype would be necessary to validate our RLCM. The main advantage of the RLCM is that it allows to spot critical links within the whole workspace much more easily and quickly than the FEM, because of the symbolic expressions of stiffness matrix elements. The RLCM is easier to use than a FEM at a pre-design stage. Once the RLCM is proved reliable enough, one can use it either to test alternative designs or choose manufacturing paths reducing the tool compliant displacement (tracking error or tracking rotational and translational compliant displacements) caused by structural compliance. Unfortunately, the FEM did not provide any results for the rotational compliance. This would be an interesting comparison since it would allow a verification of whether or not the torsional stiffness of the mobile platform obtained with the RLCM is lower compared to that of the overconstrained Orthoglide prototype described in \cite{Chablat03,ortho_web} and modeled in the FEM.
%%%%%%%%%%%%%%%%%%%%%%%%%%%%%%%%%%%%%%%%%%
%\vspace{-1cm}
\section{Conclusions}
%\vspace{-5mm}
%%%%%%%%%%%%%%%%%%%%%%%%%%%%%%%%%%%%%%%%%%
In this paper, a parametric stiffness analysis of a 3-axis PKM prototype, the Orthoglide, was conducted. First, a compliant model of the Orthoglide was obtained, then a method for parallel manipulators stiffness analysis was applied, and the stiffness matrix elements were computed symbolically in the isotropic configuration. In this configuration, the influence of the geometric design parameters on the rotational and translational stiffnesses was studied through qualitative and quantitative analysis. The analysis provided relevant and precise information for stiffness-oriented optimization of the Orthoglide.Then, the analysis of the simultaneous influence on the stiffness of two variable parameters was conducted. Such an analysis is very useful to take into account both stiffness and another performance criterion such as workspace volume or the maximal acceleration of the mobile platform. Finally, we used the symbolic expressions of the components of the compliance matrix to determine the stiffest working zone for a specific manufacturing task. The stiffest zone depends on the task and applied cutting forces. The parametric stiffness analysis shows that simple symbolic expressions carefully built and interpreted provide much information on the stiffness features of parallel manipulators, which can be relevantly used for their design and optimization.
%%%%%%%%%%%%%%%%%%%%%%%%%%%%%%%%%%%%%%%%%%
%\vspace{-5mm}

\end{document}